\documentclass{article}

\usepackage[nonatbib,preprint]{arxiv}

\usepackage[utf8]{inputenc}
\usepackage[ruled]{algorithm2e}
\usepackage{algcompatible}
\usepackage[intlimits]{amsmath}
\usepackage{titlesec}
\usepackage{float}
\usepackage{array}
\usepackage{multirow}
\usepackage[ngerman,english]{babel}
\usepackage{graphicx,wrapfig}
\usepackage[toc]{appendix}
\usepackage{etaremune}
\usepackage[inline]{enumitem}
\usepackage[onehalfspacing]{setspace}
\usepackage[T1]{fontenc}
\usepackage{amsfonts}
\usepackage{amsthm,amssymb}
\usepackage{mathtools}
\usepackage[most]{tcolorbox}
\usepackage{upgreek}
\usepackage[format=hang,font=small]{caption}
\usepackage[colorlinks,linkcolor=black,citecolor=black,urlcolor=black,breaklinks]{hyperref}
\usepackage{url}
\usepackage{booktabs}
\usepackage{nicefrac}
\usepackage{microtype}
\usepackage{fancyhdr}
\usepackage{cleveref}
\usepackage{adjustbox}
\usepackage{perpage}
\usepackage{doi}
\graphicspath{{Figs/}}
\MakePerPage{footnote}
\DeclareMathOperator*{\argmin}{arg\,min}
\DeclareMathOperator*{\argmax}{arg\,max}

\titlespacing{\section}{0pt}{5pt}{5pt}
\titlespacing{\subsection}{0pt}{5pt}{5pt}
\titlespacing{\subsubsection}{0pt}{5pt}{5pt}
\title{Active Inference-Based Optimization of Discriminative Neural Network Classifiers}
\author{Faezeh Fallah\\Institute of Signal Processing and System Theory\\University of Stuttgart,~Pfaffenwaldring 47\\70569 Stuttgart,~Germany\\\texttt{faezeh.fallah@iss.uni-stuttgart.de}}
\begin{document}
\maketitle
\begin{abstract}
Commonly used objective functions (losses) for a supervised optimization of discriminative neural network classifiers were either distribution-based or metric-based. The distribution-based losses were mostly based on the cross entropy and fitted the network model to the distribution of the training samples. This could compromise the generalization (predictive performance on unseen samples) or cause classification biases towards the dominant classes of an imbalanced class-sample distribution. The metric-based losses could make the network model independent of any distribution and thus improve its generalization. However, the metrics involved in them were binary classification metrics. This implied to decompose a multiclass classification into a series of one-vs-all classifications and then form the overall loss from an average of the one-vs-all losses. This averaging could naturally lead to a bias towards the dominant classes. Moreover, the metric-based losses could suffer from discrepancies when a class was absent in both the reference (ground truth) labels and the predicted labels. To tackle these issues, recent works have used a combination of the distribution-based and metric-based losses. In this paper, we formulated the optimization of a discriminative neural network classifier within the framework of active inference and showed that the cross entropy-based losses were indeed the variational free energy of a retrospective active inference. Then, we proposed a novel optimization process which not only tackled the unbalancedness of the class-sample distribution of the training samples but also provided a mechanism to tackle errors in the reference (ground truth) labels of the training samples. This was achieved by proposing a novel algorithm to find candidate classification labels of the training samples during the network optimization and a novel objective function for the optimizations. The algorithm could find the candidate labels of the training samples from their prior probabilities and the currently estimated posteriors on the network. The proposed objective function incorporated these candidate labels along with the original reference labels and the priors of the training samples while still being distribution-based. The proposed algorithm was the result of casting the generalized Kelly criterion for optimal betting into a multiclass classification problem. To this end, we showed that the objective function of the generalized Kelly criterion was a tight upper bound of the expected complexity of the expected free energy of a prospective active inference. This in turn allowed us to derive our proposed objective function from such an expected free energy. The incorporation of the priors into the optimization not only helped to tackle errors in the reference labels but also allowed to reduce classification biases towards the dominant classes by focusing the attention of the neural network on important but minority foreground classes.
\end{abstract}
\section{Background and Motivation}
\label{sec:BackMotivNeuralNet}

\subsection{Active Inference}
\label{ssec:activeInference}
Bayesian inference enabled perception, learning, and decision making in a passive or active perceptual task. This perception could be over a categorical (multinomial) distribution of independent and mutually exclusive states. This distribution assigned one probability to each state of each observation with the sum of these probabilities for each observation being one. That is, each observation could only be in one state at a time.
In an active perception, an agent actively engaged with its environment to gather information, seek preferred observations, avoid unpreferred observations, and take actions which could reduce uncertainty and maximize reward. If the states, observations, and policies (actions) could be discretized, then the tasks could be formulated over categorical distributions of the states, observations, and policies. These formed a discrete state-space model in which the time could be discrete as well. An active perception ruled by the Bayesian inference was called an \textbf{active inference}. The Bayesian inference inferred joint/posterior distribution of a generative/discriminative model by using the Bayes' theorem.
For the classification/segmentation tasks addressed in this dissertation, a discriminative model was sufficient. Thus, we restricted the use of the active inference to a discriminative model and only involved the posteriors in our formulations \cite{Smith2022}.

According to the Bayes' theorem, for each observation (o), state (s), and policy ($\pi$), the posterior $p(s|o,\pi)$ could be deduced from the likelihood $p(o|s,\pi)$ as
\begin{equation}
\label{eq:bayesTheorem}
p(s|o,\pi)=\frac{p(o|s,\pi)\cdot p(s|\pi)}{p(o|\pi)}
\end{equation}
with $p(o|\pi)=\sum_{s|\pi}p(o|s,\pi)$ being the model evidence or the marginal likelihood. This way, the Bayesian inference enabled perception, learning, and decision making by model inversion, i.e. deduction of the posterior $p(s|o,\pi)$ from the likelihood $p(o|s,\pi)$. This resulted in a maximum a posteriori estimation. In a simpler approach, a maximum likelihood estimation might be followed. However, the maximum likelihood estimation was prone to overfitting because the likelihoods only encoded the aleatoric uncertainty of the model caused by noise (disturbances) in its process. The epistemic (cognitive) uncertainty of the model was reflected by the states' priors ${\{p(s|\pi)\}}_s$ and the model evidence $p(o|\pi)$ included in the posteriors.
The computation of the model evidence implied to sum the likelihoods of every observation over all possible states. For most of the categorical distributions this computation was intractable. Also, by increasing the number of the states the number of the summation terms increased exponentially. For continuous distributions this summation mostly turned into a nonconvex integration of no closed-form (analytical) solution. To enable a computationally tractable active inference, the Bayes' theorem got approximated by minimizing
\begin{itemize}[leftmargin=*]
\item variational free energy (VFE)\footnote{The term \textbf{free energy} stemmed from connections between the Bayesian inference and the Bayesian mechanics ruling free energy in particular (quantum) physics elaborated by neuroscientists \cite{Friston2019}.} for perception and learning
\item expected free energy (EFE) for optimal decision making, planning, and action selection.
\end{itemize}
Each of the aforementioned objective functions depended on the policies (actions). Accordingly, the minimization of each of them provided an estimate of the posteriors conditioned on the policies. However, the VFE resulted from a course of policies based on the observations in the past and present but the EFE resulted from a course of policies based on the observations in the future. Thus, the VFE and the EFE respectively enabled retrospective and prospective policy evaluations. This difference mattered in the cases where optimal policies for the past or present were not the optimal policies for the future or vice versa.
To derive the aforementioned objectives, negative logarithm of both sides of the Bayes' formula was taken and $-\mathrm{ln}\big(p(o|\pi)\big)$ was introduced to be the self-information or surprisal\footnote{Use of the natural logarithm resulted in information being measured in \textit{nats}. In contrast, use of the $\mathrm{log}_2$ resulted in information being measured in \textit{bits}.} of the model evidence $p(o|\pi)$. Then, the VFE got defined to be the upper bound of this quantity. This way, by minimizing the VFE, the surprisal or deviation between observations and predictions of the model got minimized or the amount of evidence an observation could provide for the model got maximized, i.e. the model evidence got maximized.

As detailed in \cite{Smith2022}, the objective function of the VFE was given by
\begin{equation}
\label{eq:objFuncVFE}
\begin{split}
\mathcal{L}_{\mathrm{VFE}}&=\mathrm{KL}\Big[p(s|\pi)||q(s|\pi)\Big]-\mathrm{E}_{p(s|\pi)}\Big[\mathrm{ln}\big(q(o|s)\big)\Big]\\
&=\underbrace{\mathrm{E}_{p(s|\pi)}\Big[\mathrm{ln}\big(p(s|\pi)\big)-\mathrm{ln}\big(q(s|\pi)\big)\Big]}_{\mathrm{complexity}}-\underbrace{\mathrm{E}_{p(s|\pi)}\Big[\mathrm{ln}\big(q(o|s)\big)\Big]}_{\mathrm{accuracy}}\\
&=\sum_{s|\pi}p(s|\pi)\cdot\mathrm{ln}\big(p(s|\pi)\big)-\sum_{s|\pi}p(s|\pi)\cdot\mathrm{ln}\big(q(s|\pi)\big)-\sum_{s|\pi}p(s|\pi)\cdot\mathrm{ln}\big(q(o|s)\big)\\
&=\underbrace{\sum_{s|\pi}p(s|\pi)\cdot\mathrm{ln}\big(p(s|\pi)\big)}_{-\mathrm{entropy}}+\underbrace{\sum_{s|\pi}-p(s|\pi)\cdot\mathrm{ln}\big(q(o|\pi)\big)}_{\mathrm{cross~entropy}}
\end{split}
\end{equation}
with $q(\cdot)$ being the distribution approximating the true distribution $p(\cdot)$, $\mathrm{KL}[p(\cdot)||q(\cdot)]$ being the Kullback-Leibler (KL) divergence (dissimilarity) between $p(\cdot)$ and $q(\cdot)$, and $\mathrm{E}_{p(s|\pi)}[\cdot]$ being the expectation with respect to $p(s|\pi)$. The KL divergence was derived from the Akaike information criterion (AIC) measuring the goodness of a model in terms of its underfitting (estimation bias on seen samples) and overfitting (predictive variance on unseen samples). The AIC measured the amount of information loss (relative entropy) resulted from representing a model with another model. Here, the cross entropy was not a distance metric because the cross entropy of two identical distributions equaled their entropy. However, after subtracting the entropy from the cross entropy, the KL divergence become a distance metric. That is, the KL divergence of two identical distributions was zero \cite{Kullback1951,McMillan1956}. This way, the minimization of $\mathcal{L}_{\mathrm{VFE}}$ amounted to finding the distribution $q(\cdot)$ which best fitted $p(\cdot)$. The best fit was the minimizer of the \textbf{complexity} (overfitting) and the maximizer of the \textbf{accuracy}. The minimization of $\mathcal{L}_{\mathrm{VFE}}$ was independent of $p(s|\pi)$. Thus by adding the \textbf{entropy} term to $\mathcal{L}_{\mathrm{VFE}}$, an objective function called the cross entropy loss was obtained as
\begin{equation}
\label{eq:objFuncCrossEntr}
\mathcal{L}_{\mathrm{CE}}=-\sum_{s|\pi}p(s|\pi)\cdot\mathrm{ln}\big(q(o|\pi)\big).
\end{equation}
If $q(\cdot)$ was Gaussian, then the cross entropy loss become a sum of squared errors.

The minimization of the EFE selected optimal policies (actions) by solving the explore-exploit dilemma \cite{Friston2019}. That is, when information about the states were not enough, it emphasized on exploration (maximization of information gain or minimization of uncertainty). When the information was enough, it emphasized on exploitation (maximization of reward or minimization of expected complexity). The choice of the exploratory or the exploitative optimization depended on the current uncertainty and the future (expected) reward. This way, the minimization of the EFE sought the policies which could lead to future observations optimizing the trade-off between the maximization of the information gain and the maximization of the reward. These self-evidencing observations were called to be \textbf{preferred}. The incidence probability of a preferred observation $o$ was denoted by $p(o)$.
As detailed in \cite{Smith2022}, the objective function of the EFE was given by\\
\resizebox{1.0\textwidth}{!}{
\begin{minipage}{\textwidth}
\begin{subequations}
\label{eq:objFuncEFE}
\begin{align*}
\mathcal{L}_{\mathrm{EFE}}&=\mathrm{KL}\Big[p(o)||q(o|\pi)\Big]+\mathrm{E}_{p(s|\pi)}\Big[\mathrm{H}\big[q(o|\pi)\big]\Big]\tag{\ref{eq:objFuncEFE}}\\
&=\underbrace{\mathrm{E}_{p(o)}\Big[\mathrm{ln}\big(p(o)\big)-\mathrm{ln}\big(q(o|\pi)\big)\Big]}_{\mathrm{expected~complexity}}+\underbrace{\mathrm{E}_{p(s|\pi)}\Big[\mathrm{H}\big[q(o|\pi)\big]\Big]}_{\mathrm{uncertainty}}\\
&=\underbrace{\sum_{o}p(o)\cdot\Big[\mathrm{ln}\big(p(o)\big)-\mathrm{ln}\big(q(o|\pi)\big)\Big]}_{\mathrm{expected~complexity}}+\underbrace{\sum_{s|\pi}-p(s|\pi)\cdot\sum_{o|\pi}q(o|\pi)\cdot\mathrm{ln}\big(q(o|\pi)\big)}_{\mathrm{uncertainty}}
\end{align*}
\end{subequations}
\vspace{1mm}
\end{minipage}}
with $\mathrm{H}\big[q(o|\pi)\big]=-\sum_{o|\pi}q(o|\pi)\cdot\mathrm{ln}\big(q(o|\pi)\big)$ being the entropy of $q(o|\pi)$. This way, active inference provided a unified mathematical framework to model interdependent aspects of perception, learning, and decision making. This framework could build highly flexible and generalizable generative models which could explain neuro-cognitive behavioral processes as well as partially observable Markov decision processes \cite{Friston2019,Smith2022}.

\subsection{Optimization of Discriminative Neural Network Classifiers}
\label{ssec:optNeuralNetClass}
A neural network was composed of several perceptrons (nodes) in multiple layers. The layers included an input layer, some hidden layers, and an output layer. A perceptron contained a nonlinear function called an activation and was connected to other perceptrons in neighboring layers via some weights and a bias. These weights, biases, and the nonlinear activations formed \textbf{main parameters} of the neural network. Besides, the neural network had some \textbf{hyperparameters} defining its architecture and its optimization process.
Neural networks have demonstrated promising results in a wide range of applications. This was due to the \textbf{universal approximation theorem} stating that a feed-forward network with a hidden layer containing a finite number of neurons (perceptrons) could approximate any continuous function on a compact subset of $\mathbb{R}^d$ if and only if the used activations (perceptrons' nonlinearities) were nonpolynomial. The number of the parameters of such an approximating model defined its capacity to represent and to predict patterns. For a fully connected neural network, this number was $\mathcal{O}(n_{\mathrm{layer}}\cdot n_{\mathrm{width}}^2)$ where $n_{\mathrm{layer}}$ was the number of layers (depth of the network) and $n_{\mathrm{width}}^2$ was the number of perceptrons per layer (width of the network). Thus, an increase in the width increased the number of the parameters faster than an increase in the number of layers. An increase in the number of parameters increased the chance of overfitting. Moreover, a wide shallow network could fit to the patterns in the seen (training) samples but could not predict the patterns in unseen (validation or test) samples. To enhance the generalization (predictive performance on unseen samples), the neural network should contain more layers (become deeper) \cite{Dean2012,Ruder2016,Goodfellow2016}.

In a fully connected neural network, every perceptron was connected to all the perceptrons in its neighboring layers. This network lacked the capability of capturing regional (intra-layer) neighborhood patterns and thus needed handcrafted features to accomplish its task. To have an end-to-end neural network, directly applicable to the input samples without any preprocessing or explicit feature extraction, the features should be extracted by the network itself. This implied to capture regional (intra-layer) neighborhood patterns through limited receptive fields. The receptive field of a perceptron defined the size and the shape of the region at the input of the network affecting the output of the perceptron. The receptive field was determined by the kernel and the dept of the perceptron in the neural network. The deeper the perceptron in the network was the larger its receptive field become.

The application of a perceptron's kernel to its inputs returned a number of feature maps. By increasing the receptive field of the perceptron, the number and the abstraction level of its feature maps got increased but the size of each map got decreased. Accordingly, by using different kernels and locating the perceptrons at different depths of the network, features of different resolutions and abstraction levels could be obtained. Besides capturing subtle features and patterns, a kernel-based network enabled \textbf{weight sharing} by applying the same kernel coefficients to various regions in space. This resulted in a significantly lower number of parameters than a fully connected network and thus reduced the chance of overfitting and improved the generalization (predictive performance on unseen samples). In addition, it reduced the number of samples needed to train (optimize) the network. An easy-to-implement kernel for estimating a categorical distribution in a classification problem or a continuous distribution in a regression task was \textbf{convolutional}\footnote{In practice, many machine learning libraries avoided the \textit{sign flip} action involved in the convolution and thus simply implemented a cross correlation between the inputs and the kernels of each layer.}. This type of kernel formed a convolutional neural network (CNN) which could be end-to-end and deep as well.

\begin{figure}[t!]
\begin{center}
\includegraphics[width=1.0\textwidth]{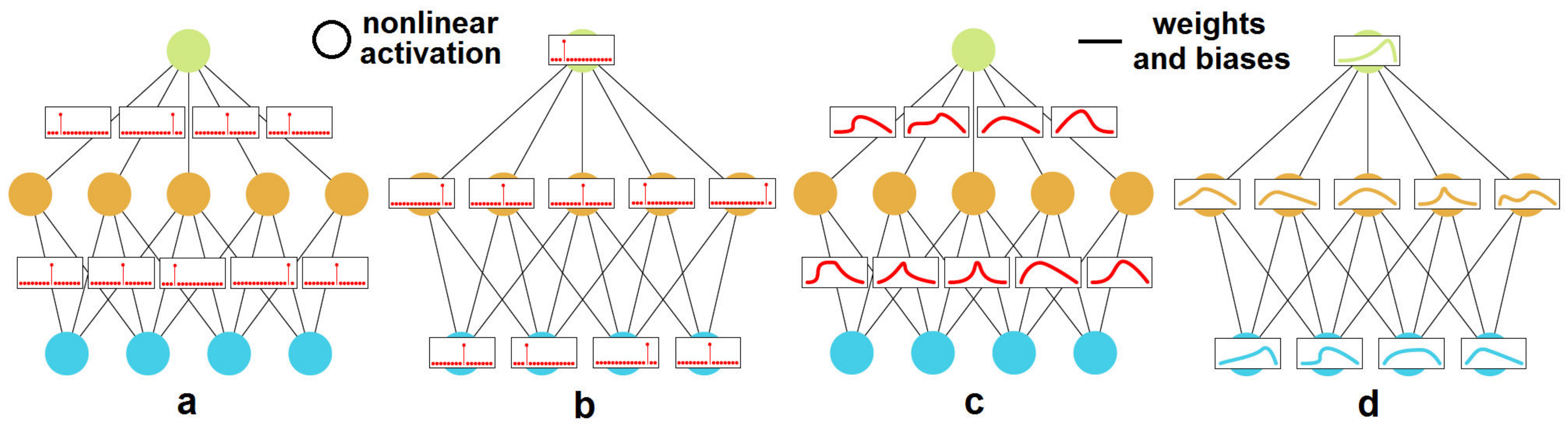}
\caption{A neural network with plain (single-valued) weights and biases (a), plain activations (b), Bayesian (distributed) weights and biases (c), and Bayesian activations (d).}
\label{fig:PlainBaysNetwork}
\end{center}
\end{figure}

As shown in \autoref{fig:PlainBaysNetwork}, a neural network could be \textbf{plain} or \textbf{Bayesian}. In the plain network, each parameter, i.e. each weight, bias, or activation, had a single value. In the Bayesian network, each parameter had a vector of values representing its distribution and uncertainty. The Bayesian network was formed from an ensemble of plain networks. That is, multiple plain networks got built and then the Bayesian network's parameters got derived from a weighted average of the plain networks' parameters with the weight of each network being the posteriors estimated by it for the training samples. Accordingly, whatever derived or concluded for the plain networks could be extended to the Bayesian networks. In the following, we simply referred to the plain neural network as the neural network. Such a network demanded an objective function and a process to optimize its parameters as well as a regularization to mitigate overfitting. A commonly used objective function for such a network was the cross entropy loss introduced in \eqref{eq:crossEntrLoss}. The commonly used optimization processes were based on the gradient (first derivative) descent of the objective function \cite{Kingma2015}. The regularization was mostly done by penalizing large perceptrons' weights or dropping perceptrons of low confident weights in a method called Dropout \cite{Gal2015,Jospin2022}.

The gradient descent optimization relied on the fact that the opposite direction of the gradient (first derivative) of the scalar field of the objective function pointed to the minimum of the function. Accordingly, in each iteration $i\in\{1,\cdots,n_{\mathrm{it}}\}$ of this optimization, a movement in the direction of the negative gradient of the objective function at the current point updated the network's parameters. This optimization had a linear complexity with regard to the number of network's parameters. The gradient at each iteration was the average gradient of the training samples passed through the network's layers. The samples could be passed one-by-one or all at once. The former led to a stochastic and the latter led to a batch-based optimization. A complete pass through all the training samples was called an \textit{epoch} \cite{Dean2012,Ruder2016,Goodfellow2016}.

The averaging of the gradients of the batch's samples resulted in a smooth variation of the cost versus the iterations. In addition, the batch-based optimization allowed to apply vectorized and parallelized operations. However, it was restricted to convex or relatively smooth error manifolds and could only find local minima. Moreover, feeding a large batch of samples become memory intensive. The stochastic gradient descent optimization updated the network's parameters by passing one sample through the network in each iteration. This could avoid memory issues, could address nonconvex optimizations, and could even find global minima. However, due to a more frequent update of the network's parameters it resulted in fluctuating cost versus the iterations. Depending on the samples' gradients the fluctuations might never reach a minimum but rather dance around it. Moreover, the stochastic optimization could not benefit from the vectorized or the parallelized operations.

An intermediate between the stochastic and the batch-based optimization was a mini-batch-based optimization. In this approach, the training samples got divided into $n_{\mathrm{batch}}$ disjoint batches, i.e. $\mathbb{T}_{\mathrm{train}}=\cup_{b=1}^{n_{\mathrm{batch}}}\mathbb{T}_b$. Then, in each iteration $i\in\{1,\cdots,n_{\mathrm{it}}\}$, the samples of one batch got passed through the network and the average gradient of these samples updated the network's parameters. The size or the number of the batches was a hyperparameter. This way, by adapting the size or the number of the batches, the mini-batch-based optimization could utilize the vectorized and the parallelizable operations to speed up its computations while fitting the fluctuations of the cost versus the iterations to the nonconvexity of the addressed problem. Accordingly, if $n_{\mathrm{epoch}}$ was the number of epochs, then the network was optimized by $n_{\mathrm{it}}=(|\mathbb{T}_{\mathrm{train}}|/|\mathbb{T}_b|)\times n_{\mathrm{epoch}}$ iterations. In each epoch, the batches and the samples of each batch got randomly shuffled to avoid overfitting to some of the samples.

With $\alpha_{\mathrm{lr}}\in(0,1)$ being the learning rate (step size), $\boldsymbol{\eta}^{(i)}$ being the vector of the main parameters of the neural network in the iteration $i\in\{1,\cdots,n_{\mathrm{it}}\}$, and $\nabla_{\boldsymbol{\eta}^{(i)}}(\mathcal{L})$ being the gradient of a generic objective function $\mathcal{L}$ with regard to these parameters, we had
\begin{equation}
\label{eq:updateMainParams}
\boldsymbol{\eta}^{(i)}=\boldsymbol{\eta}^{(i-1)}-\alpha_{\mathrm{lr}}\cdot\boldsymbol{\delta}^{(i)}.
\end{equation}

In the gradient descent optimization, $\boldsymbol{\delta}^{(i)}=\nabla_{\boldsymbol{\eta}^{(i-1)}}(\mathcal{L})$. This resulted in a slow convergence and sensitivity to abrupt variations of the gradient due to noise and perturbations. To speed up the convergence, to propel out of local minima, and to smooth out the gradient variations, in the method of \textit{momentum}, $\boldsymbol{\delta}^{(i)}$ got defined to be an exponentially weighted moving average (first moment) of the current and past gradients. The averaging weight was a decay rate called first moment rate $\beta_{\mathrm{fm}}\in[0,1)$. It emphasized the importance of recent gradients to the older ones. For $\beta_{\mathrm{fm}}=0$, the momentum boiled down to the gradient descent. For $\beta_{\mathrm{fm}}=1$ and $\alpha_{\mathrm{lr}}\approx 0$ it resulted in endless fluctuations of the cost versus the iterations like the movements of a ball in a frictionless bowl. Two major bottlenecks of the gradient descent and the momentum were the possibility of being trapped into saddle points (i.e. points of zero gradients in all directions) and a slow update in the directions of sparse features of weak gradients. To tackle these, the adaptive gradient algorithm (AdaGrad) defined $\boldsymbol{\delta}^{(i)}$ to be the instant (current) gradient divided (normalized) by the square root of the sum of the squared gradients. This scaling allowed to avoid saddle points and adapted the gradient and thus the optimization rate in each direction to its history of updates. That is, the more a feature (direction) was updated in the past the less it would be updated in the future.

Despite of these improves, the AdaGrad was slow since the sum of the squared gradients only grew but never shrank. This growth also resulted in a rapid decay of $\boldsymbol{\delta}^{(i)}$ and thus a poor performance in dealing with nonconvex objective functions and dense features (directions of strong gradients). The root mean square propagation (RMSprop) fixed these issues by replacing the sum of the squared gradients with an exponentially weighted moving average of the squared gradients. This was called second moment of the gradient. The averaging weight was a decay rate called the second moment rate $\beta_{\mathrm{sm}}\in[0,1)$. It emphasized the importance of recent gradients to the older ones. Moreover, in the formation of $\boldsymbol{\delta}^{(i)}$, the division (normalization) of the instant gradient by the second moment balanced the step size. More specifically, it decreased the step size for large gradients to prevent their explosion and increased the step size for small gradients to prevent their vanishing. The exploding and the vanishing gradients were common issues of deep neural networks.

The adaptive moment estimation (Adam) combined the momentum (first moment) with the RMSprop (second moment) to take advantages of both. This was done by defining the $\boldsymbol{\delta}^{(i)}$ to be the first moment divided (normalized) by the second moment. This way, the Adam got the convergence speed from the momentum and the ability to adapt the gradients in different directions from the RMSprop \cite{Kingma2015}. More specifically,
\begin{align}
\label{eq:adamDelta}
\nonumber
&\boldsymbol{\delta}^{(i)}=\hat{\boldsymbol{m}}^{(i)}\oslash\big(\sqrt[\circ]{\hat{\boldsymbol{v}}^{(i)}}\oplus{10^{-8}}\big)~~~&\mathbf{g}^{(i)}=\nabla_{\boldsymbol{\eta}^{(i-1)}}(\mathcal{L})\\\nonumber
&\text{biased first moment:}~~&\boldsymbol{m}^{(i)}=\beta_{\mathrm{fm}}\odot\boldsymbol{m}^{(i-1)}\oplus(1-\beta_{\mathrm{fm}})\odot\mathbf{g}^{(i)}\\
&\text{bias-corrected first moment:}~~&\hat{\boldsymbol{m}}^{(i)}=\boldsymbol{m}^{(i)}\oslash(1-\beta_{\mathrm{fm}}^i)\\\nonumber
&\text{biased second moment:}~~&\boldsymbol{v}^{(i)}=\beta_{\mathrm{sm}}\odot\boldsymbol{v}^{(i-1)}\oplus(1-\beta_{\mathrm{sm}})\odot\mathbf{g}^{(i)}\odot\mathbf{g}^{(i)}\\\nonumber
&\text{bias-corrected second moment:}~~&\hat{\boldsymbol{v}}^{(i)}=\boldsymbol{v}^{(i)}\oslash(1-\beta_{\mathrm{sm}}^i).
\end{align}

All the aforementioned techniques relied on the gradient (first derivative) of the scalar field of the objective function of the neural network. The second derivative of this scalar field was represented by a Hessian matrix. Commonly used optimization techniques based on the Hessian matrix were the Newton and the quasi-Newton method, the conjugate gradient method, and the Levenberg-Marquardt algorithm \cite{Dean2012,Ruder2016}.
A common way to optimize a network's parameters by any one of the derivative-based techniques was a backpropagation. This method demanded the objective function to be expressed in terms of the network's outputs (goodness of the model) and to be differentiable with respect to the outputs of every layer. In case of using the gradient of the objective function with respect to the network's parameters, this gradient got expressed as a product of the layerwise errors. Then, the backpropagation took the following steps:
\begin{itemize}[leftmargin=*]
\item initialized the network's parameters with random numbers.
\item passed a batch through all the layers and computed the outputs of every layer.
\item computed the error at the last layer by comparing the predictions with the references.
\item propagated the error from the last layer to the first layer to find the error of each layer.
\item expressed the gradient of the objective function as a product of the layerwise errors.
\item updated the network's parameters according to \eqref{eq:updateMainParams}.
\end{itemize}

\subsection{Commonly Used Objective Functions}
\label{ssec:CommObjsBotlncks}
For a probabilistic estimate, the outputs of the neural network got converted to probabilities (posteriors) by using a softmax (normalized exponential) function. This function converted a vector to another vector whose elements summed up to one and each element of the output had a monotonic relationship with an element of the input. In our case, the input vector was the network's outputs for each sample and had a length of $n_{\mathrm{clas}}=|\mathbb{L}|$. This way, the output of the softmax function could be interpreted as a categorical probability distribution of a multinomial classification over $n_{\mathrm{clas}}$ mutually exclusive classes. That is, every sample could only have one reference classification label. A special case of the softmax function was the sigmoid function. This function assumed that the classes were independent but not mutually exclusive. Thus, every sample could have multiple reference labels. The sigmoid function cast a multinomial classification into a series of binary (one-vs-all) classifications. Accordingly, its outputs did not necessarily sum up to one. For a sample $v_{b,j}\in\mathbb{T}_b\subseteq\mathbb{T}_{\mathrm{train}}$, the network's outputs at the $i^{\mathrm{th}}$ iteration of the optimization formed a vector $\mathbf{z}_{b,j}^{(i)}={[z_{b,j,c}^{(i)}]}_{c\in\mathbb{L}}$. Then, the posteriors $\hat{\mathbf{p}}_{b,j}^{(i)}={[\hat{p}_{b,j,c}^{(i)}]}_{c\in\mathbb{L}}$ produced by applying the softmax function to these outputs were
\begin{equation}
\label{eq:softmaxPostNet}
\hat{p}_{b,j,c}^{(i)}=\frac{\mathrm{exp}\big(z_{b,j,c}^{(i)}\big)}{\sum_{k\in\mathbb{L}}{\mathrm{exp}\big(z_{b,j,k}^{(i)}\big)}}\in(0,1)~~~~~\text{with}~~~~~\sum_{c\in\mathbb{L}}{\hat{p}_{b,j,c}^{(i)}}=1.
\end{equation}
Accordingly, if the training samples $\mathbb{T}_b\subseteq\mathbb{T}_{\mathrm{train}}$ were used to optimize the network's parameters in the iteration $i\in\{1,\cdots,n_{\mathrm{it}}\}$, then $\mathbf{L}_b={[\mathbf{l}_{b,j}]}_j={[\mathbf{l}_{b,c}]}_c={[l_{b,j,c}]}_{j,c}$ was the $|\mathbb{T}_b|\times n_{\mathrm{clas}}$ matrix of vectorized reference labels of these samples, $\mathbf{Z}_b^{(i)}={[\mathbf{z}_{b,j}^{(i)}]}_j={[z_{b,j,c}^{(i)}]}_{j,c}$ was the $|\mathbb{T}_b|\times n_{\mathrm{clas}}$ matrix of the network's outputs for these samples, and $\hat{\mathbf{P}}_b^{(i)}={[\hat{\mathbf{p}}_{b,j}^{(i)}]}_j={[\hat{p}_{b,j,c}^{(i)}]}_{j,c}$ was the $|\mathbb{T}_b|\times n_{\mathrm{clas}}$ matrix of their classification posteriors estimated by the network.

If the reference (ground truth) labels of the training samples $\mathbb{T}_{\mathrm{train}}$ were provided at the time of optimization (training), then for each sample $v_{b,j}\in\mathbb{T}_b\subseteq\mathbb{T}_{\mathrm{train}}$ the vector $\mathbf{l}_{b,j}$ was a one-hot-encoding of its reference label $l_{b,j}\in\mathbb{L}$ and was given by
\begin{equation}
\label{eq:refLabelOneHot}
\mathbf{l}_{b,j}={[l_{b,j,c}]}_{c\in\mathbb{L}}~~~~~\text{with}~~~~~l_{b,j,c}=\begin{cases}1&\text{if}~c=l_{b,j}=\text{reference~label~of~}v_{b,j}\in\mathbb{T}_b\\0&\text{otherwise}\end{cases}.
\end{equation}
If the reference (ground truth) labels of the training samples $\mathbb{T}_{\mathrm{train}}$ were not provided at the time of optimization (training), then for each sample $v_{b,j}\in\mathbb{T}_b\subseteq\mathbb{T}_{\mathrm{train}}$ the vector $\mathbf{l}_{b,j}$ was
\begin{equation}
\label{eq:refLabelUniform}
\mathbf{l}_{b,j}={[l_{b,j,c}]}_{c\in\mathbb{L}}=\frac{1}{n_{\mathrm{clas}}}\odot\mathbf{1}_{n_{\mathrm{clas}}=|\mathbb{L}|}.
\end{equation}

For a discriminative neural network classifier acting on $|\mathbb{L}|=n_{\mathrm{clas}}$ classes, a common way to evaluate the estimated posteriors against the reference labels was to use the cross entropy loss introduced in \eqref{eq:objFuncCrossEntr}. In this application, the policies $\pi$ incorporated in \eqref{eq:objFuncCrossEntr} represented the network's parameters. Each state $s$ was a class $c\in\mathbb{L}$ and each observation $o$ was a sample $v_{b,j}\in\mathbb{T}_b\subseteq\mathbb{T}_{\mathrm{train}}$. Accordingly, $p(s|\pi)=p(s)$ was the occurrence probability of a class (state) $s$ which could be represented by the vectorized reference labels of the samples (observations). Also, $q(o|\pi)$ was the classification posterior estimated by the network's parameters $\pi$ for the reference classification label of a sample (observation) $o$. With these, the cross entropy loss of the discriminative neural network classifier become
\begin{equation}
\label{eq:crossEntrLoss}
\mathcal{L}_{\mathrm{CE}}(\hat{\mathbf{P}}_b^{(i)},\mathbf{L}_b)=\frac{-1}{|\mathbb{L}|\cdot|\mathbb{T}_b|}\sum_{j\in\mathbb{T}_b}\sum_{c\in\mathbb{L}}l_{b,j,c}\cdot\mathrm{ln}\big(\hat{p}_{b,j,c}^{(i)}\big).
\end{equation}
If the posteriors were generated by the softmax function, then this loss was called a \textbf{softmax cross entropy loss}. As detailed in \eqref{eq:objFuncVFE}, the cross entropy loss resulted from the minimization of the VFE through minimizing the KL divergence (dissimilarity) between the reference distribution $p(\cdot)$ and the estimated distribution $q(\cdot)$. In a categorical classification, the reference distribution $p(\cdot)$ was the histogram of the class-sample distribution of the training samples. The estimated distribution $q(\cdot)$ was a known function parametrized with the network's parameters. This way, the cross entropy loss and the objective functions of the active inference compared the distributions and thus were \textbf{distribution-based}.
If the class-sample distribution of the training samples was imbalanced, then it had maxima at the dominant classes. These maxima formed minima of the cross entropy loss. Thus, any minimizer of the cross entropy loss could be trapped into those minima and could thus return classifications biased towards the dominant classes of the training samples.

To reduce the impacts of the dominant classes on the optimization of a neural network, the cross entropy loss got weighted and/or modulated. The resulting losses included
\begin{enumerate}[label={(\arabic*)},leftmargin=*]
\item\textbf{weighted cross entropy loss} which weighted the contribution of each class $c\in\mathbb{L}$ by the inverse of its frequency $w_{b,c}\in(0,1)$ in the batch $\mathbb{T}_b\subseteq\mathbb{T}_{\mathrm{train}}$ and (optionally) weighted the contribution of each sample $v_{b,j}\in\mathbb{T}_b\subseteq\mathbb{T}_{\mathrm{train}}$ by its distance $d_{b,j,1}\in\mathbb{R}_{\geq 0}$ to the border of the nearest class and its distance $d_{b,j,2}\in\mathbb{R}_{\geq 0}$ to the border of the second nearest class through the weight $w_{b,j}\in(0,1)$ \cite{Ronneberger2015,Badrinarayanan2016}
\begin{equation}
\label{eq:WCE}
\mathcal{L}_{\mathrm{WCE}}(\hat{\mathbf{P}}_b^{(i)},\mathbf{L}_b)=\frac{-1}{|\mathbb{L}|\cdot|\mathbb{T}_b|}\sum_{j\in\mathbb{T}_b}\sum_{c\in\mathbb{L}}w_{b,j,c}\cdot l_{b,j,c}\cdot\mathrm{ln}\big(\hat{p}_{b,j,c}^{(i)}\big)
\end{equation}
\begin{equation}
\label{eq:wghtdCrssEntrWght}
w_{b,j,c}=w_{b,c}+w_{b,j}=\underbrace{\frac{\sum_{k\in\mathbb{L}}|\mathbb{T}_{b,k}|}{|\mathbb{T}_{b,c}|+{10^{-8}}}}_{w_{b,c}\in(0,1)}+\underbrace{w_{\mathrm{mo}}\cdot\mathrm{exp}\Big(-\frac{(d_{b,j,1}+d_{b,j,2})^2}{2\cdot\sigma_{\mathrm{mo}}^2}\Big)}_{w_{b,j}\in(0,1)}
\end{equation}
with $w_{\mathrm{mo}}=10$, $\sigma_{\mathrm{mo}}=5$, and $|\mathbb{T}_{b,c}|=\mathrm{card}\Big(\{l_{b,j,c}=1\}\Big)$. The distances to the classification borders could be computed by applying morphological operators to the samples in the classification domain, e.g. the spatial domain in an image segmentation task.
\item\textbf{focal (modulated cross entropy) loss} which weighted the contribution of each class by the difficulty of classifying its samples with the difficulties being highlighted with a modulation factor $\gamma_{\mathrm{mod}}\in\mathbb{R}_{+}$. That is, the higher the $\gamma_{\mathrm{mod}}\in\mathbb{R}_{+}$ was, the more the easy samples got downweighted to emphasize the role of the difficult samples \cite{Lin2018}
\begin{equation}
\label{eq:focalLoss}
\mathcal{L}_{\mathrm{FL}}(\hat{\mathbf{P}}_b^{(i)},\mathbf{L}_b)=\frac{-1}{|\mathbb{L}|\cdot|\mathbb{T}_b|}\sum_{j\in\mathbb{T}_b}\sum_{c\in\mathbb{L}}{\big(1-\hat{p}_{b,j,c}^{(i)}\big)}^{\gamma_{\mathrm{mod}}}\cdot l_{b,j,c}\cdot\mathrm{ln}\big(\hat{p}_{b,j,c}^{(i)}\big).
\end{equation}
\item\textbf{weighted focal loss} which additionally weighted the contribution of each class $c\in\mathbb{L}$ by the inverse of its frequency $w_{b,c}\in(0,1)$ in the batch $\mathbb{T}_b\subseteq\mathbb{T}_{\mathrm{train}}$ \cite{Lin2018}
\begin{equation}
\label{eq:wghtFocLoss}
\mathcal{L}_{\mathrm{WFL}}(\hat{\mathbf{P}}_b^{(i)},\mathbf{L}_b)=\frac{-1}{|\mathbb{L}|\cdot|\mathbb{T}_b|}\sum_{j\in\mathbb{T}_b}\sum_{c\in\mathbb{L}}w_{b,c}\cdot{\big(1-\hat{p}_{b,j,c}^{(i)}\big)}^{\gamma_{\mathrm{mod}}}\cdot l_{b,j,c}\cdot\mathrm{ln}\big(\hat{p}_{b,j,c}^{(i)}\big).
\end{equation}
\end{enumerate}

The weighted cross entropy and the weighted focal loss highlighted the role of the minority classes over the role of the majority classes by including the weight $w_{b,c}\in(0,1)$ in their terms. This way, the more a class had training samples, the less its classification errors contributed to the overall loss. In a so-called class-balanced cross entropy loss \cite{Cui2019}, each weight $w_{b,c}\in(0,1)$ got defined based on the \textit{effective number} $n_{b,c}\in(0,1)$ of the training samples of the class $c\in\mathbb{L}$ in the feature space as
\begin{equation}
\label{eq:classBalWght2}
w_{b,c}=\bigg[1-\frac{n_{b,c}-1}{n_{b,c}}\bigg]/\bigg[1-\Big(\frac{n_{b,c}-1}{n_{b,c}}\Big)^{|\mathbb{T}_{b,c}|}\bigg].
\end{equation}
This method assumed that each sample in the feature space covered a subspace and the overall samples' subspaces of each class formed its prototypical subspace. Then, the volume of this prototype defined the effective number of the class. However, in most of the applications, the feature space was hardly accessible. In a neural network, it was also variable across the network's layers. Moreover, the computation of the subspace coverages in the feature space was expensive and depending on the dimensionality and the geometry of the space. Accordingly, in \cite{Cui2019}, each number $n_{b,c}\in(0,1)$ got handled as a hyperparameter.

The aforementioned weighting and modulation schemes could reduce the impacts of the dominant classes of the seen (training) samples on the network's optimization. However, they were still based on the cross entropy loss and thus fitted the network's model to the seen distribution. This could compromise the network's generalization (predictive performance on unseen samples) when the distribution of the unseen (validation or test) samples differed from the distribution of the seen (training) samples. An objective evaluation of a classifier on unseen samples could be done through several metrics. Among these metrics, the Dice coefficient (DICE) and its equivalent the Jaccard index (JI) provided perceptual clues, scale invariance, and counts of false positive and false negative mispredictions. The JI was also called the intersection over union (IoU) and the DICE was the F-$\beta$ score with $\beta=1$. These metrics could be computed with a low complexity. This enabled their integration into an iterative optimization of neural network classifiers in the form of \textbf{metric-based} losses. Then, the optimum network's parameters were the \textbf{maximizers} of the DICE \cite{Milletari2016} or the \textbf{minimizers} of the Jaccard distance (JD)=1$-$JI=1$-$IoU \cite{Bertels2019}.

The DICE=F-1 score and the JD=1$-$JI=1$-$IoU directly compared the binary masks of the predicted and the reference labels of the training samples without considering their distribution. This made the network's model independent of any distribution and could thus tackle the differences of the seen and unseen distributions. However, the binary masks compared by these metrics got formed from discrete-valued labels. This hindered to integrate those metrics into a continuous optimizer with backpropagation. More specifically, the predicted labels were the results of applying an \textbf{arg~max} operation to the classification posteriors $\hat{\mathbf{p}}_{b,j}^{(i)}={[\hat{p}_{b,j,c}^{(i)}]}_{c\in\mathbb{L}}$ estimated by the network. This operation was nonlinear, irreversible, and indifferentiable. Thus, to integrate the metrics into a continuous optimizer with backpropagation, the network's outputs $\mathbf{z}_{b,j}^{(i)}={[z_{b,j,c}^{(i)}]}_{c\in\mathbb{L}}$ should be stored in each iteration $i\in\{1,\cdots,n_{\mathrm{it}}\}$ and for each sample $v_{b,j}\in\mathbb{T}_b\subseteq\mathbb{T}_{\mathrm{train}}$. These storages got retrieved during the backpropagation and thus increased the memory footprint of the network and hindered to optimize a large network with a large number of samples per batch \cite{Bertels2019}.

To integrate the aforementioned metrics into a continuous optimization framework, they should be replaced by their continuous relaxed (real-valued) surrogates. For the DICE, this surrogate compared the vectorized reference labels $\mathbf{L}_b={[\mathbf{l}_{b,j}]}_j={[\mathbf{l}_{b,c}]}_c={[l_{b,j,c}]}_{j,c}$ against the classification posteriors $\hat{\mathbf{P}}_b^{(i)}={[\hat{\mathbf{p}}_{b,j}^{(i)}]}_j={[\hat{p}_{b,j,c}^{(i)}]}_{j,c}$ estimated by the network as
\begin{equation}
\label{eq:DICEloss}
\mathcal{L}_{\mathrm{DICE}}(\hat{\mathbf{P}}_b^{(i)},\mathbf{L}_b)=\frac{2}{|\mathbb{L}|}\sum_{c\in\mathbb{L}}\frac{\sum_{j\in\mathbb{T}_b}l_{b,j,c}\cdot\hat{p}_{b,j,c}^{(i)}}{\sum_{j\in\mathbb{T}_b}\big[l_{b,j,c}^2+\hat{p}_{b,j,c}^{{(i)}^2}\big]}.
\end{equation}

The above DICE loss was reversible and differentiable and could thus be integrated into a gradient descent optimization with backpropagation \cite{Milletari2016}. However, its \textbf{nonconvexity} hindered its wide use in many applications. Other metrics such as the mean symmetric surface distance and the Hausdorff distance were also nonconvex besides being too complex for an iterative optimization process \cite{Jadon2020}. In addition, each discrete-valued metric was a set function mapping from a set of mispredictions to a set of real numbers. However, among them, only the set function of the JD was submodular. This allowed to find a convex closure of the JD in a polynomial time. This convex closure was a \textbf{convex continuous} relaxed (real-valued) surrogate taking nonnegative real-valued mispredictions as inputs. Another metric of these properties was the Hamming distance. The convex closure of the JD got derived according to the smooth convex Lov\'{a}sz extension of submodular set functions \cite{Berman2018,Bertels2019}. The JD was defined as
\begin{equation}
\label{eq:JD}
\text{Jaccard distance (JD)=1$-$JI=}\resizebox{0.64\hsize}{!}{%
$\frac{|\mathbb{V}_{\mathrm{prd}}\cup\mathbb{V}_{\mathrm{ref}}|\setminus|\mathbb{V}_{\mathrm{prd}}\cap\mathbb{V}_{\mathrm{ref}}|}{|\mathbb{V}_{\mathrm{prd}}\cup\mathbb{V}_{\mathrm{ref}}|}=\frac{|\mathbb{V}_{\mathrm{prd}}\setminus\mathbb{V}_{\mathrm{ref}}|+|\mathbb{V}_{\mathrm{ref}}\setminus\mathbb{V}_{\mathrm{prd}}|}{|\mathbb{V}_{\mathrm{prd}}\cup\mathbb{V}_{\mathrm{ref}}|}.$}
\end{equation}
Based on this definition, the set function of the JD for the batch $\mathbb{T}_b\subseteq\mathbb{T}_{\mathrm{train}}$ and the class $c\in\mathbb{L}$ in the iteration $i\in\{1,\cdots,n_{\mathrm{it}}\}$ was
\begin{subequations}
\label{eq:setFuncJD}
\begin{equation}
\text{JD}:~~~\mathbb{M}_{b,c}^{(i)}\in\{0,1\}^{|\mathbb{T}_b|}\longmapsto\frac{\mathrm{nnz}\Big(\mathbb{M}_{b,c}^{(i)}\Big)}{\mathrm{nnz}\Big(\{l_{b,j,c}=1\}\cup\{\hat{l}_{b,j,c}^{(i)}=1\}\Big)}\in\mathbb{R}
\end{equation}
\begin{equation}
\text{with}~~~\hat{l}_{b,j,c}^{(i)}=\begin{cases}1&\text{if}~c=\argmax_k\{\hat{p}_{b,j,k}^{(i)}\}\\0&\text{otherwise}\end{cases}~~~\text{forming}~~~\hat{\mathbf{l}}_{b,j}^{(i)}={[\hat{l}_{b,j,c}^{(i)}]}_{c\in\mathbb{L}}
\end{equation}
\begin{equation}
\text{and}~~~\mathbb{M}_{b,c}^{(i)}=\Big[\big\{l_{b,j,c}=1,\hat{l}_{b,j,c}^{(i)}\neq1\big\}\cup\big\{l_{b,j,c}\neq1,\hat{l}_{b,j,c}^{(i)}=1\big\}\Big]\in\{0,1\}^{|\mathbb{T}_b|}
\end{equation}
\end{subequations}
being the set of mispredictions defined over the discrete hypercube $\{0,1\}^{|\mathbb{T}_b|}$. Also, $\mathrm{nnz}(\mathbb{M}_{b,c}^{(i)})$ was the number of nonzero elements of the binary set $\mathbb{M}_{b,c}^{(i)}$. To form the convex continuous surrogate of the JD, first $\mathbb{M}_{b,c}^{(i)}\in\{0,1\}^{|\mathbb{T}_b|}$ should be replaced by a nonnegative real-valued misprediction vector $\mathbf{m}_{b,c}^{(i)}={[m_{b,j,c}^{(i)}]}_j\in\mathbb{R}_{\geq 0}^{|\mathbb{T}_b|}$. Then, the surrogate should be found in $\mathbb{R}_{\geq 0}^{|\mathbb{T}_b|}$. This search was NP-hard unless the JD was submodular. According to Proposition 11 in \cite{Yu2020}, the set function $\text{JD}:\{0,1\}^{|\mathbb{T}_b|}\longmapsto\mathbb{R}$ was submodular. That is,
\begin{equation}
\label{eq:submodular}
\forall\mathbb{M}_1,\mathbb{M}_2\in\{0,1\}^{|\mathbb{T}_b|}:~~~\text{JD}(\mathbb{M}_1)+\text{JD}(\mathbb{M}_2)\geq\text{JD}(\mathbb{M}_1\cup\mathbb{M}_2)+\text{JD}(\mathbb{M}_1\cap\mathbb{M}_2).
\end{equation}
Under this condition, the convex closure of $\text{JD}:\{0,1\}^{|\mathbb{T}_b|}\longmapsto\mathbb{R}$ in $\mathbb{R}_{\geq 0}^{|\mathbb{T}_b|}$ was tight and continuous and could be computed in a polynomial time. This convex closure was called the Lov\'{a}sz extension and was given in \cite{Lovasz1983,Fujishige1991} as
\begin{equation}
\label{eq:lovaszJacDis}
\begin{split}
\overline{\text{JD}}:&~~~\mathbf{m}_{b,c}^{(i)}\in\mathbb{R}_{\geq 0}^{|\mathbb{T}_b|}\longmapsto\bigg[\frac{1}{|\mathbb{T}_b|}\sum_{j\in\mathbb{T}_b}m_{b,j,c}^{(i)}\cdot{g_j\big(\mathbf{m}_{b,c}^{(i)}\big)}\bigg]\in\mathbb{R}\\
\text{with}&~~~g_j\big(\mathbf{m}_{b,c}^{(i)}\big)=\text{JD}\big(\{u_1,\cdots,u_j\}\big)-\text{JD}\big(\{u_1,\cdots,u_{j-1}\}\big)
\end{split}
\end{equation}
being the $j^{\mathrm{th}}$ element of the gradient $\mathbf{g}\big(\mathbf{m}_{b,c}^{(i)}\big)$ and $\{u_1,\cdots,u_{|\mathbb{T}_b|}\}$ denoting a permutation of the elements of $\mathbf{m}_{b,c}^{(i)}={[m_{b,j,c}^{(i)}]}_j$ in descending order, i.e. ${[\mathbf{m}_{b,c}^{(i)}]}_{u_1}\geq\cdots\geq{[\mathbf{m}_{b,c}^{(i)}]}_{u_{|\mathbb{T}_b|}}$. Thus, the $\overline{\text{JD}}\big(\mathbf{m}_{b,c}^{(i)}\big)$ was a weighted average of the elements of the misprediction vector $\mathbf{m}_{b,c}^{(i)}\in\mathbb{R}_{\geq 0}^{|\mathbb{T}_b|}$ with the weights being the elements of the first derivative (gradient) of $\overline{\text{JD}}$ with respect to $\mathbf{m}_{b,c}^{(i)}\in\mathbb{R}_{\geq 0}^{|\mathbb{T}_b|}$. This way, the Lov\'{a}sz extension $\overline{\text{JD}}$ interpolated $\text{JD}$ in $\mathbb{R}_{\geq 0}^{|\mathbb{T}_b|}\setminus\{0,1\}^{|\mathbb{T}_b|}$ while having the same values as $\text{JD}$ on $\{0,1\}^{|\mathbb{T}_b|}$ \cite{Bach2013,Berman2018}.

For a binary classification, the misprediction vector $\mathbf{m}_{b,c}^{(i)}={[m_{b,j,c}^{(i)}]}_j\in\mathbb{R}_{\geq 0}^{|\mathbb{T}_b|}$ was given by $m_{b,j,c}^{(i)}=\mathrm{max}\big[(1-z_{b,j,c}^{(i)}\cdot l_{b,j,c}),~0\big]$ with $\mathbf{z}_{b,j}^{(i)}={[z_{b,j,c}^{(i)}]}_{c\in\mathbb{L}}$ being the network's outputs (before the softmax function) at the $i^{\mathrm{th}}$ iteration for the sample $v_{b,j}\in\mathbb{T}_b\subseteq\mathbb{T}_{\mathrm{train}}$. This misprediction vector resulted in a convex piecewise linear surrogate called the Lov\'{a}sz hinge loss \cite{Yu2020}.

For a multiclass classification, the misprediction vector $\mathbf{m}_{b,c}^{(i)}={[m_{b,j,c}^{(i)}]}_j\in\mathbb{R}_{\geq 0}^{|\mathbb{T}_b|}$ was formed from the classification posteriors $\hat{\mathbf{p}}_{b,j}^{(i)}={[\hat{p}_{b,j,c}^{(i)}]}_{c\in\mathbb{L}}$ produced by the softmax function in \eqref{eq:softmaxPostNet}. This misprediction vector resulted in a convex continuous surrogate with regard to the batch $\mathbb{T}_b\subseteq\mathbb{T}_{\mathrm{train}}$ and the class $c\in\mathbb{L}$ in the iteration $i\in\{1,\cdots,n_{\mathrm{it}}\}$. Thus, for the classification over $n_{\mathrm{clas}}=|\mathbb{L}|$ classes, the overall loss was an average of these class-specific surrogates. This overall loss was called the Lov\'{a}sz-Softmax loss and was given in \cite{Berman2018} as
\begin{equation}
\label{eq:lovaszSoftMult}
\begin{gathered}
\mathcal{L}_{\mathrm{LS}}(\hat{\mathbf{P}}_b^{(i)},\mathbf{L}_b)=\frac{1}{|\mathbb{L}|\cdot|\mathbb{T}_b|}\sum_{c\in\mathbb{L}}\sum_{j\in\mathbb{T}_b}m_{b,j,c}^{(i)}\cdot{g_j\big(\mathbf{m}_{b,c}^{(i)}\big)}\\
\text{with}~~~\mathbf{m}_{b,c}^{(i)}={[m_{b,j,c}^{(i)}]}_j\in\mathbb{R}_{\geq 0}^{|\mathbb{T}_b|}~~~\text{and}~~~m_{b,j,c}^{(i)}=\begin{cases}1-\hat{p}_{b,j,c}^{(i)}&\text{if}~c=l_{b,j,c}\\\hat{p}_{b,j,c}^{(i)}&\text{otherwise}\end{cases}~\in(0,1).
\end{gathered}
\end{equation}

The computation of the Lov\'{a}sz extension $\overline{\text{JD}}$ in \eqref{eq:lovaszJacDis} implied to sort the elements of $\mathbf{m}_{b,c}^{(i)}={[m_{b,j,c}^{(i)}]}_j\in\mathbb{R}_{\geq 0}^{|\mathbb{T}_b|}$ and to call the JD with the permutation order. The sort had a complexity of $\mathcal{O}\big(|\mathbb{T}_b|\cdot\mathrm{log}(|\mathbb{T}_b|)\big)$ and the call had a complexity of $\mathcal{O}(|\mathbb{T}_b|)$. However, by keeping a track of the cumulative number of false positive and false negative mispredictions, the complexity of the call could be amortized to $\mathcal{O}(1)$. That is, in each iteration, instead of computing the gradient from scratch only the gradient got updated. In this case, the overall complexity of computing \eqref{eq:lovaszJacDis} become $\mathcal{O}\big(|\mathbb{T}_b|\cdot\mathrm{log}(|\mathbb{T}_b|)\big)$. The procedure of computing the gradient of the Lov\'{a}sz-Softmax loss in \eqref{eq:lovaszSoftMult} was given by Algorithm 1 in \cite{Berman2018}.

The convexity and the differentiability of the Lov\'{a}sz-Softmax loss in \eqref{eq:lovaszSoftMult} allowed to use it as an objective function for optimizing a discriminative neural network classifier by a gradient descent optimizer with backpropagation. Also, the operations involved in its computation were differentiable and implementable on graphics processing units (GPUs).

\subsection{Baseline Architecture}
\label{ssec:baselineArchi}
Each convolutional layer of a neural network could extract features of a certain resolution while being capable of downsampling or reducing the spatial resolution by using an appropriate stride. These allowed to learn hierarchical (multiresolution) features by cascading multiple convolutional layers. The opposite of a convolutional layer was a transposed convolutional or a deconvolutional layer of similar feature learning capability but an inherent upsampling or increase of the spatial resolution. By following the convolutional layers with the deconvolutional layers an \textbf{encoder-decoder} architecture was obtained. The encoder was a downsampler, a compressor, or a contractor performing \textbf{analysis}. The decoder was an upsampler, a decompressor, or an expander performing \textbf{synthesis}. Each encoder/decoder was composed of multiple stages. Each \textbf{stage} processed features of a certain resolution through one or more convolutional/deconvolutional layers and then downsampled/upsampled its newly computed features to the next resolution. To avoid loss of information due to the downsampling, in each encoder stage, the number of the newly computed features got multiplied by the downsampling rate. Conversely, in each decoder stage, the number of the newly computed features got divided by the upsampling rate.

\begin{figure}[t!]
\begin{center}
\includegraphics[width=1.0\textwidth]{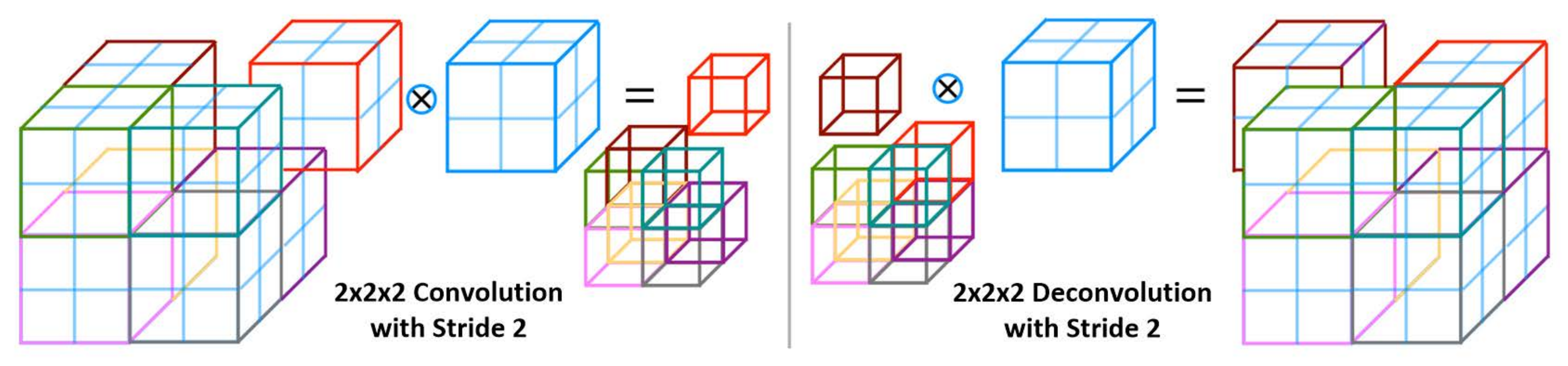}
\caption{Downsampling (left) and upsampling (right) in the V-net.}
\label{fig:VnetUpDownSampl}
\end{center}
\end{figure}

A widely used neural network of such an encoder-decoder architecture was the U-net. As the inputs passed through its encoder stages, the progressively expanding receptive fields of its convolutional layers increased the abstraction and the context of its extracted features. Thus, at the end of the encoder or bottom of the \textbf{U}, features of minimum resolution but maximum abstraction and context were obtained. The spatial resolution of these features got reconstructed by passing them through the deconvolutional layers of the decoder stages and combining them with original higher resolution features. The original features were directly obtained from the corresponding encoder stage through a skip connection. That is, features extracted by each encoder stage got forwarded to the corresponding decoder stage to compensate information loss due to the downsampling. This feature forwarding could enhance the delineation of boundaries between different classes and sped up the convergence of the optimization. At the end of the decoder, the resulting feature maps had a resolution and size like the input of the network. A weighted average of these feature maps combined them into the desired number of classes. This was done by passing them through a convolutional layer of $1\times 1\times 1$ kernel size, $0$ padding, and stride of $1$ in each dimension. As given by \eqref{eq:softmaxPostNet}, the resulting network's outputs got then passed through a softmax function to produce the estimated classification posteriors for the samples \cite{Ronneberger2015}.

\begin{table}[t!]
\begin{center}
\caption{The receptive fields and the sizes of the feature maps at different stages of the V-net.}
\vspace{2mm}
\resizebox{0.7\textwidth}{!}{%
\begin{tabular}{c|cc|c}
\multirow{2}{*}{\textbf{Stage}}&\multicolumn{2}{c|}{\textbf{Receptive Field}}&\multirow{2}{*}{\textbf{Size of Feature Maps}}\\
&\textbf{Encoder}&\textbf{Decoder}&\\
\hline
$1$&$5\times 5\times 5$&$551\times 551\times 551$&$128\times352\times256$\\
$2$&$22\times22\times22$&$546\times546\times546$&$64\times176\times128$\\
$3$&$72\times72\times72$&$528\times528\times 528$&$32\times88\times64$\\
$4$&$172\times172\times172$&$476\times476\times476$&$16\times44\times32$\\
$5$&$372\times372\times372$&$372\times372\times372$&$8\times22\times16$\\
\end{tabular}}
\label{table:recFieldFeatMap}
\end{center}
\end{table}

The downsampling and the upsampling of the U-net made it a hierarchical architecture capable of capturing, analyzing, and synthesizing features at different spatial resolutions. This way, the U-net could automatically extract local and contextual patterns. The local patterns got captured by the shallower layers and the contextual patterns by the deeper layers of a larger receptive field. At the end, the decoder synthesized (gathered and assembled) the local (high resolution) and the contextual (low resolution) features into the final classification. These enabled a localization as well as an accurate classification in any domain of any size and thus made the U-net a breakthrough for end-to-end optimizations. Moreover, making all the operations of the U-net 3D allowed to apply it to 3D volumetric domains. The 3D U-net got enhanced by making its encoder stages \textbf{residual}. That is, the input of each encoder stage got added to its output. This could mitigate vanishing gradients and speed up the convergence of the optimization \cite{He2016}. In addition, the 3D U-net could learn 3D volumetric structures out of sparsely annotated 2D slices. This allowed to use it in a semi-automated annotation process as well as a fully automated 3D detection \cite{Cicek2016,Rakhlin2018}.

In the 3D U-net, each downsampling/upsampling had a factor of 2 and was done through a max-pooling/unpoolig over a $2\times 2\times 2$ kernel with a stride of $2$ in each dimension. Also, each convolutional layer applied $0$ padding. Thus, the valid part of each feature map at the output of each convolutional layer had a smaller size than its input feature map. In addition, the 3D U-net learned the residual functions only in its encoder stages. In a so-called V-net, the 3D U-net become \textbf{fully convolutional} by applying each downsampling/upsampling through a convolutional/deconvolutional layer of a kernel size of $2\times 2\times 2$, a $0$ padding, and a stride of $2$ in each dimension. To avoid loss of information, each downsampling doubled the number of feature maps. Conversely, each upsampling halved the number of feature maps. \autoref{fig:VnetUpDownSampl} shows the downsampling and the upsampling in the V-net.

In contrast to the max-pooling/unpoolig operations, the convolution/deconvolution-based downsampling/upsampling was reversible and differentiable. These allowed to backpropagate each downsampling/upsampling without needing to store its inputs per sample and iteration. This way, the memory footprint of the V-net become much less than the 3D U-net while the analysis and comprehension of its internal process got simplified. Moreover, each convolution of the V-net applied an appropriate padding to make the feature maps at its output of the same size as its input. Furthermore, the V-net learned the residual functions not only in the encoder stages but also in the decoder stages. This further boosted its performance and sped up its optimization \cite{Milletari2016}. This way, the 3D U-net or the V-net got widely used in many applications \cite{Rakhlin2018,Sijia2022}. Accordingly, we resorted to an end-to-end optimization of the 3D fully convolutional and residual V-net for our implementations and evaluations. For this, we tailored the number and the sizes of the feature maps and the kernels of the convolutional/deconvolutional layers to our volumetric fat-water images. Also, through the network, we processed the data in an N$\times$D$\times$H$\times$W$\times$C format with N=$|\mathbb{T}_b|$ being the number of the volumetric fat-water images in each batch, C being the number of the feature maps, D being the depth, H being the height, and W being the width of each feature map. We trained (optimized) the V-net by using a mini-batch-based gradient descent optimizer with backpropagation and a sufficiently large input volume to capture as much contextual information as possible. Due to the memory limitations of the used GPU\label{numBatches}, we could only include 2 volumetric fat-water images in each batch. Moreover, each volumetric fat-water image had 2 channels containing its voxelwise fat and water intensities. Accordingly, at the input of the network, N$\times$D$\times$H$\times$W$\times$C=$2\times128\times352\times256\times2$.

Each encoder/decoder stage of the V-net extracted and learned features of a certain spatial resolution by using one to three 3D (volumetric) convolutional/deconvolutional layers. In our case, each of these layers had a kernel size of $5\times 5\times 5$, a padding of $2$, and a stride of $1$ in each dimension. Also, regarding the size of our images and the sizes of the addressed objects (tissues) in our segmentations, we found \textbf{5 stages (resolution levels)} to be sufficient for our hierarchical feature learning. \autoref{table:recFieldFeatMap} shows the receptive fields and the sizes of the feature maps at different stages. As can be seen, the innermost (deepest) stage of the network could already capture the entire context of the input volume. This allowed to perceive the whole anatomy of interest and ensured access to enough contextual information for reliably classifying each voxel at the output of the neural network classifier.

\begin{figure}[t!]
\begin{center}
\includegraphics[width=1.0\textwidth]{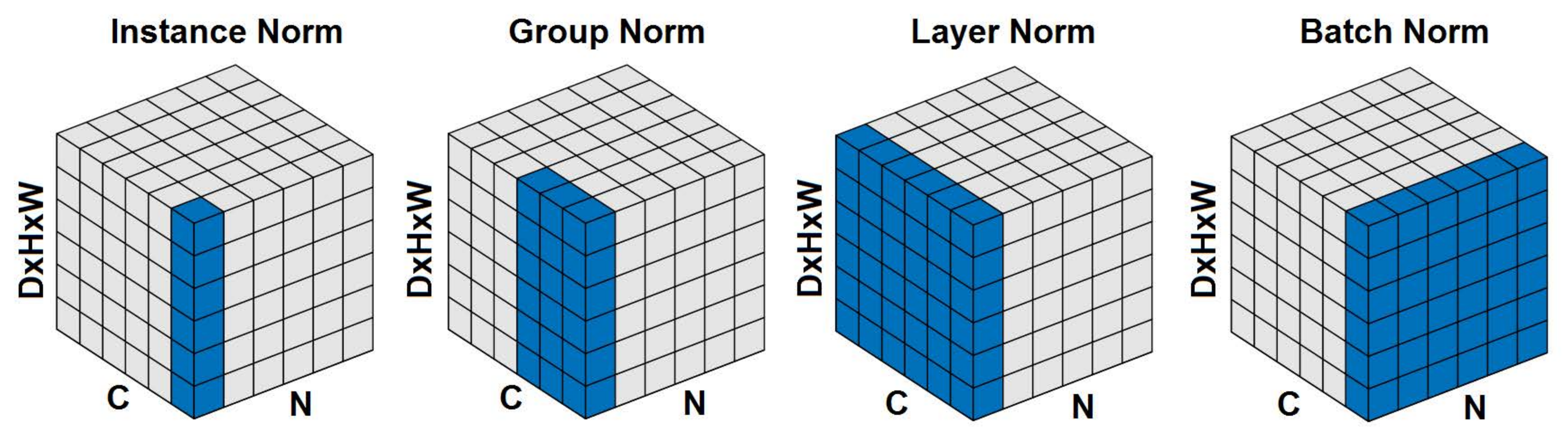}
\caption{Different normalization techniques applied to a feature map of size N$\times$D$\times$H$\times$W$\times$C with N denoting the number of batches, C denoting the number of channels, and D$\times$H$\times$W denoting the spatial dimensions. In each case, the blue voxels got normalized by the same mean and variance aggregated across them.}
\label{fig:NormTechs}
\end{center}
\end{figure}

Besides the convolutional/deconvolutional layers, each \textbf{residual} encoder/decoder stage normalized its feature maps and applied nonlinearities to them. Like the original V-net, we used a parametric rectified linear unit (PReLU) with a parameter $a_{\mathrm{prelu}}\in\mathbb{R}_{\geq 0}$ for each nonlinear activation. The parameter $a_{\mathrm{prelu}}\in\mathbb{R}_{\geq 0}$ controlled the outputs for negative inputs and thus was called the coefficient of leakage. It got optimized along with the main parameters (weights and biases) of the network. The normalization of the feature maps decoupled the lengths of the network's gradients from their directions. This could accelerate the convergence of the optimizations and thus allowed higher learning rates. It could also stabilize the optimizations by mitigating the internal covariate shift\footnote{changes of stochastic distributions of the inputs of each layer of the network due to the changes of the parameters of the previous layers}, enhancing the robustness against the initializations, and smoothing the objective function. Moreover, it could penalize large network's weights and thereby reduce the overfitting or improve the generalization.

We modified the V-net by changing the type of the normalization from batch normalization \cite{Ioffe2015} to instance (contrast) normalization \cite{Ulyanov2016}. The commonly used batch normalization was based on mini-batch statistics. That is, during the training, the mean and the variance of each feature map of each batch got learned across all the dimensions (D, H, W) and all the N members of the batch to normalize (remove bias and scale of) the corresponding feature map in the evaluation phase. The instance normalization took a similar approach. However, it computed the mean and the variance of each feature map of each batch only across the dimensions (D, H, W). In case of having a small batch size, like our case, the exponential moving averages of the mean and the variance of each feature map of each batch had strong fluctuations across the training iterations. This was due to the poor statistical power of the small batch and thereby made the batch normalization ineffective. In this case, the instance normalization was more effective and consistent \cite{Ulyanov2016}\label{instanceNorm}. Other varieties of the normalization were the layer and the group normalization \cite{Wu2020}. \autoref{fig:NormTechs} shows their differences to the batch and the instance normalization.

\begin{figure}[t!]
\begin{center}
\includegraphics[width=0.8\textwidth]{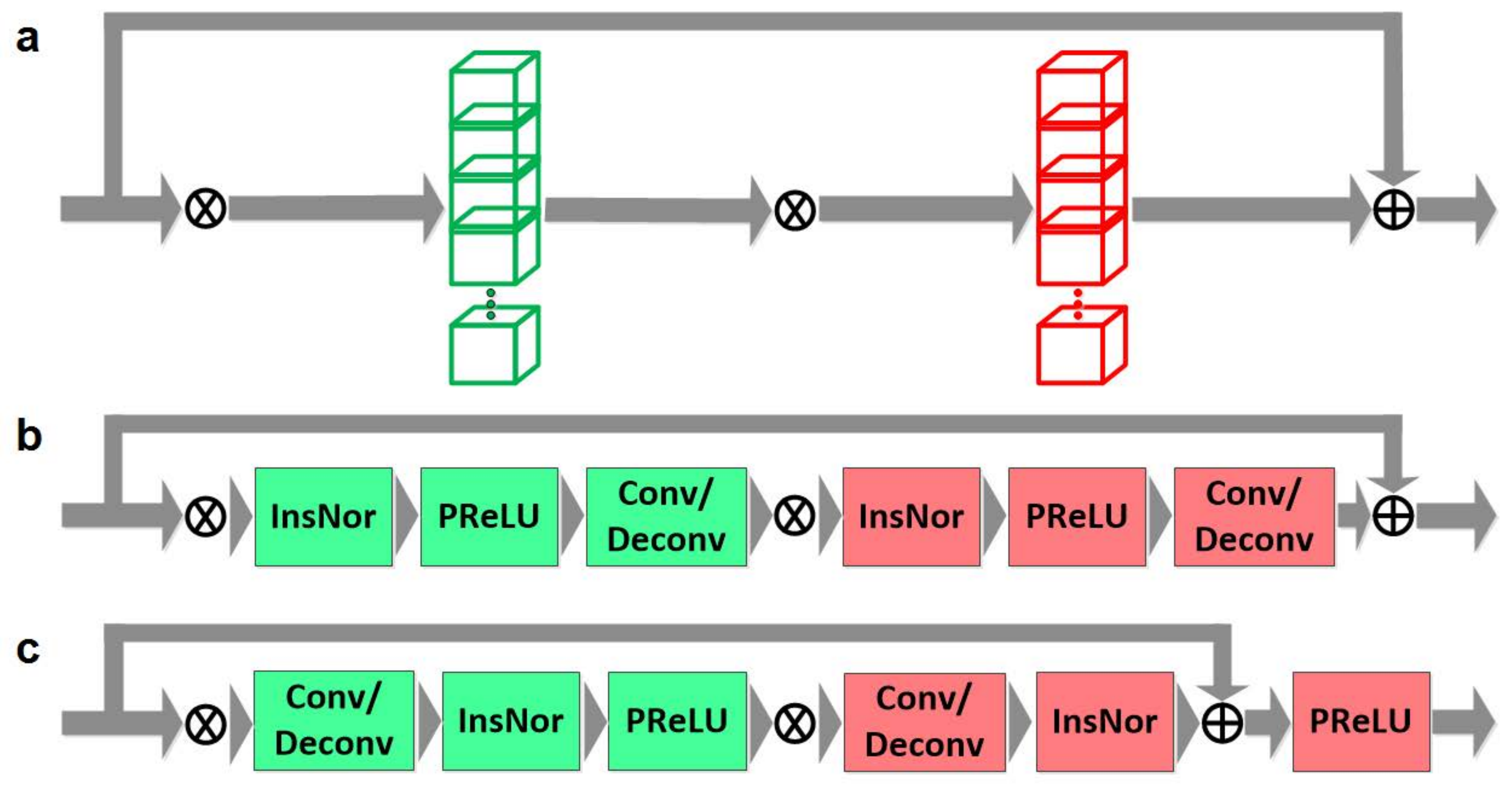}
\caption{A residual encoder/decoder stage comprising 2 convolutional/deconvolutional layers (a) with the new (b) and the original order (c) of the operations.}
\label{fig:ResEncDec}
\end{center}
\end{figure}

\begin{figure}[t!]
\begin{center}
\includegraphics[width=0.97\textwidth,height=19.9cm]{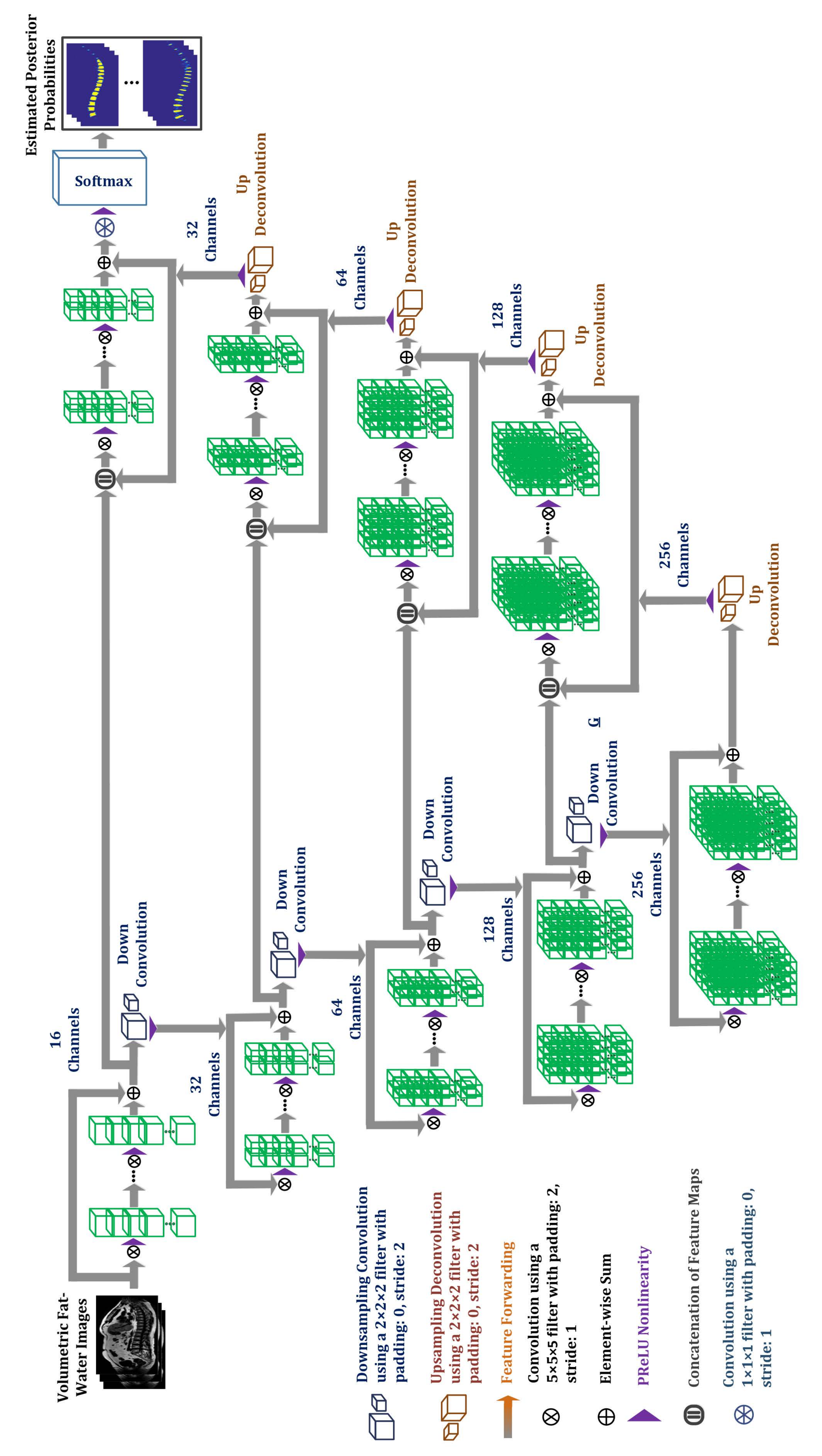}
\caption{Schematic of the 3D fully convolutional and residual V-net with the encoder and the decoder stages on its left and right side, respectively.}
\label{fig:Vnet}
\end{center}
\end{figure}

We also modified the V-net by changing the order of operations in each \textbf{residual} encoder/decoder stage. Instead of the convention of applying the normalization between the convolution/deconvolution and the nonlinear activation, as suggested in \cite{Kaiming2016}, we applied a full preactivation normalization and removed after-addition activation. \autoref{fig:ResEncDec} compares the new and the original orders of the operations of a residual encoder/decoder stage comprising 2 convolutional/deconvolutional layers. The advantage of the new order was that it made the overall nonlinear function of each stage a real identity mapping. This enabled a direct and clean propagation of signals from one stage to another stage in both forward and backward directions. Other kinds of skip connections which involved a sort of scaling (like the Dropout), gating, or convolution/deconvolution on the signal path could hamper a clean propagation of the information and thus lead to optimization problems. Moreover, the new order could improve the generalization of the network's model by reducing its overfitting. That is, it increased the error on seen (training) samples but reduced the error on unseen (validation or test) samples. Furthermore, in the original order, addition of the shortcut to the normalized signal made the overall signal at the input of the last nonlinear activation unnormalized. However, in the new order, the signal at the input of each nonlinear activation was normalized. \autoref{fig:Vnet} shows the described V-net architecture.

To mitigate \textbf{overfitting} and the \textbf{imbalanced class-sample distribution} of the training samples, \textbf{attention mechanisms} got proposed. These methods aimed to focus the attention of the network's parameters on important (foreground) minority classes. This attention could reduce the training samples to an \textbf{effective subset} of a lower unbalancedness than the original set. It could also vanish the redundant or irrelevant network's parameters by suppressing feature activations in irrelevant regions of the classification domain. These in turn reduced the overfitting and sped up the convergence of the network's optimization. The attention could be stimulated by \textbf{incorporating priors} into the optimization process and/or \textbf{modifying the network's architecture}. Neither the cross entropy-based nor the metric-based losses, defined in \autoref{ssec:CommObjsBotlncks}, could accommodate the priors of the samples. Consequently, the attention mechanisms were restricted to architectural modifications.

Trainable (optimizable) attention mechanisms were categorized as \textbf{hard} or \textbf{soft}. The hard attention mechanisms iteratively cropped a region of interest through a Monte Carlo sampling optimized by a reinforcement learning. These sampling-based updates were indifferentiable and thus hard to optimize. The soft attention mechanisms involved a differentiable model composed of real-valued parameters. Thus, they could be optimized through a gradient descent optimizer with backpropagation. The output of the soft attention model for each feature map was a probabilistic map called \textbf{attention map}. In an additive or a multiplicative attention mechanism this map got computed by adding or multiplying the filtered feature map(s) by a filtered gating map, respectively. If the attention map was commuted by a convolutional neural network (CNN), then each filter was a convolutional layer. The attention mechanism turned into a self-attention if the gating maps were produced internally. The elementwise multiplication or addition of each attention map with its corresponding feature map highlighted salient features for the classification. This enabled an attention-based \textbf{feature pooling or pruning}. If the gating maps brought contextual information, then the feature pooling was with regard to the contextual dependencies of the features. Besides mitigating the overfitting and the imbalanced class-sample distribution of the training samples, the attention-based feature pooling could enhance the sensitivity, the prediction accuracy, and the robustness of the neural network classifier. A commonly used architecture for soft attention was a region proposing feed-forward CNN. A bottleneck of this approach was its excessive and redundant use of the model's parameters and features. This could increase the overall optimization overhead and the overfitting before the convergence of the optimization could realize any attention for a possible reduction of the network's parameters \cite{Oktay2018}.

As mentioned earlier, the U-net and the V-net were capable of extracting (analyzing) and reconstructing (synthesizing) multiresolution (multiscale) features. This was done by extracting coarser features through downsampling the feature maps across the encoder stages and then reconstructing finer (higher resolution) features across the decoder stages. To this end, the receptive field at the coarsest resolution was to be large enough to capture all the contextual information highlighting the overall category and location of the foreground classes. After the localization, the finer (higher resolution) features delineated boundaries between different classes more precisely. These altogether allowed to capture large shape and size variations in the classification domain and thus improved the classification accuracy.

The reconstruction of the finer (higher resolution) features in each decoder stage was with the help of the features extracted by the corresponding encoder stage at the same spatial resolution. This feature forwarding reduced redundant and repeated computation of the features and thus enhanced efficiency in the usage of the computational power and memory. The plain skip connection of the feature forwarding path could be replaced by an \textbf{attention gate} realizing an \textbf{attention-based feature pooling}. This pooling vanished redundant features right before the concatenation of the original features with the reconstructed features. This way, it could suppress irrelevant regions in the classification domain by vanishing redundant network's perceptrons. This in turn reduced the overfitting of the network and the unbalancedness of the samples' distribution seen at the time of its training (optimization). Furthermore, the computational overhead of such an attention gate was much lower than the region proposing CNN. This and the reduction of the network's parameters could reduce the computational complexity of the optimizations and speed up their convergence \cite{Oktay2018}.

A promising self-attention mechanism for integration into each feature forwarding path of the U-net or the V-net was a grid-based gating module. In this approach, each gating map was not fixed across the elements of its corresponding feature maps for which the attention maps were to be computed. Instead, it was a feature map of a lower (coarser) resolution already generated by the network itself. This way, the resulting attention maps were grid-based (i.e. variable across the elements of the feature maps) and could thus highlight salient features with respect to local patterns. The gating based on the feature maps of a lower (coarser) resolution allowed to consider a bigger context in the feature pooling and thereby disambiguated irrelevant and noisy features. Moreover, the grid-based gating module eliminated the need to an external explicit region proposing CNN by implicitly proposing soft (probabilistic) map of the target structures on the fly. This attention mechanism could be trained from scratch to focus on the target structures of varying shapes and sizes without additional supervision. Its filters (linear transformations) downweighted the gradients from irrelevant regions and could thus be implemented through convolutional layers filtering the network's activations in both forward and backward passes \cite{Oktay2018}.

\begin{figure}[t!]
\begin{center}
\includegraphics[width=1.0\textwidth]{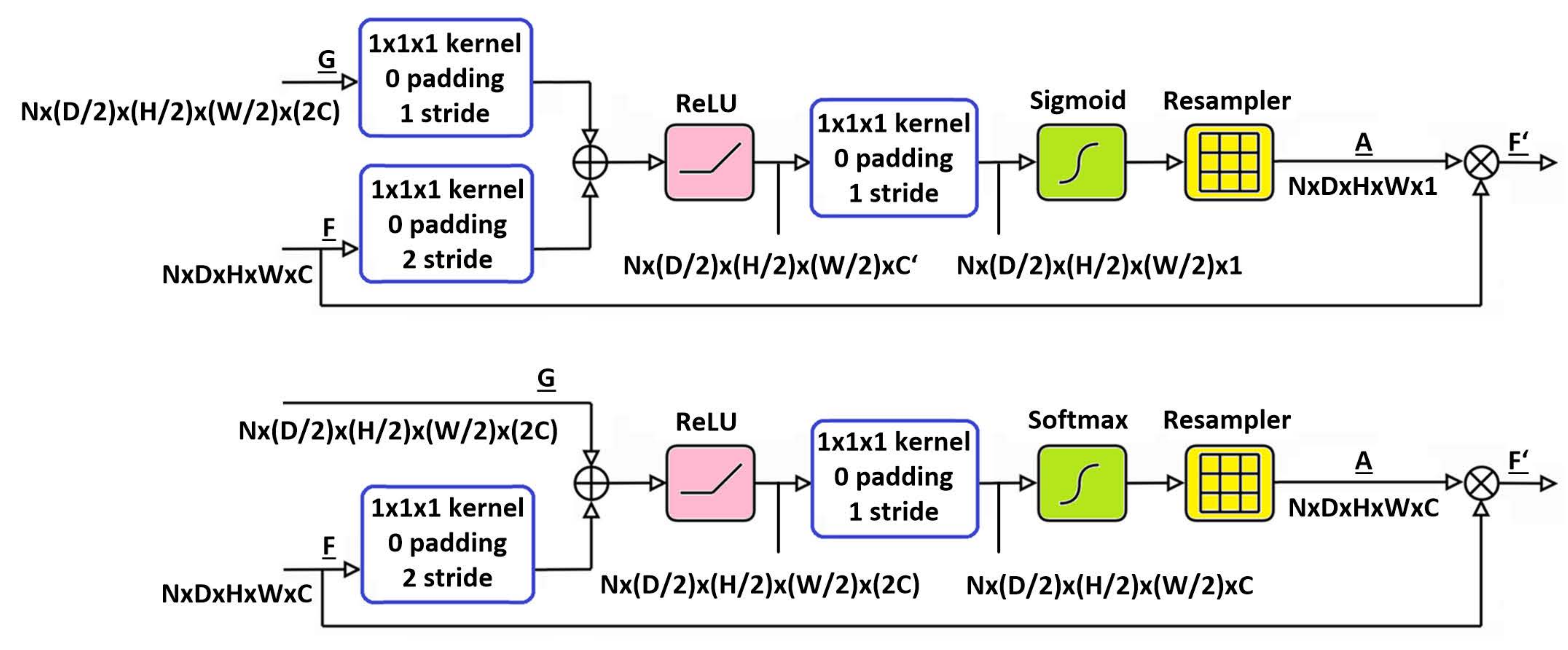}
\caption{Schematic of the original (\textbf{upper row}) and the proposed (\textbf{lower row}) grid-based gating module with $\underline{\mathbf{F}}$ denoting the tensor of the input feature maps, $\underline{\mathbf{G}}$ denoting the tensor of the gating maps, $\underline{\mathbf{A}}$ denoting the tensor of the attention maps, $\underline{\mathbf{F}}'$ denoting the tensor of the output feature maps, and each blue box depicting a convolutional layer.}
\label{fig:AttentionGate}
\end{center}
\end{figure}

\begin{figure}[t!]
\begin{center}
\includegraphics[width=0.95\textwidth,height=19.5cm]{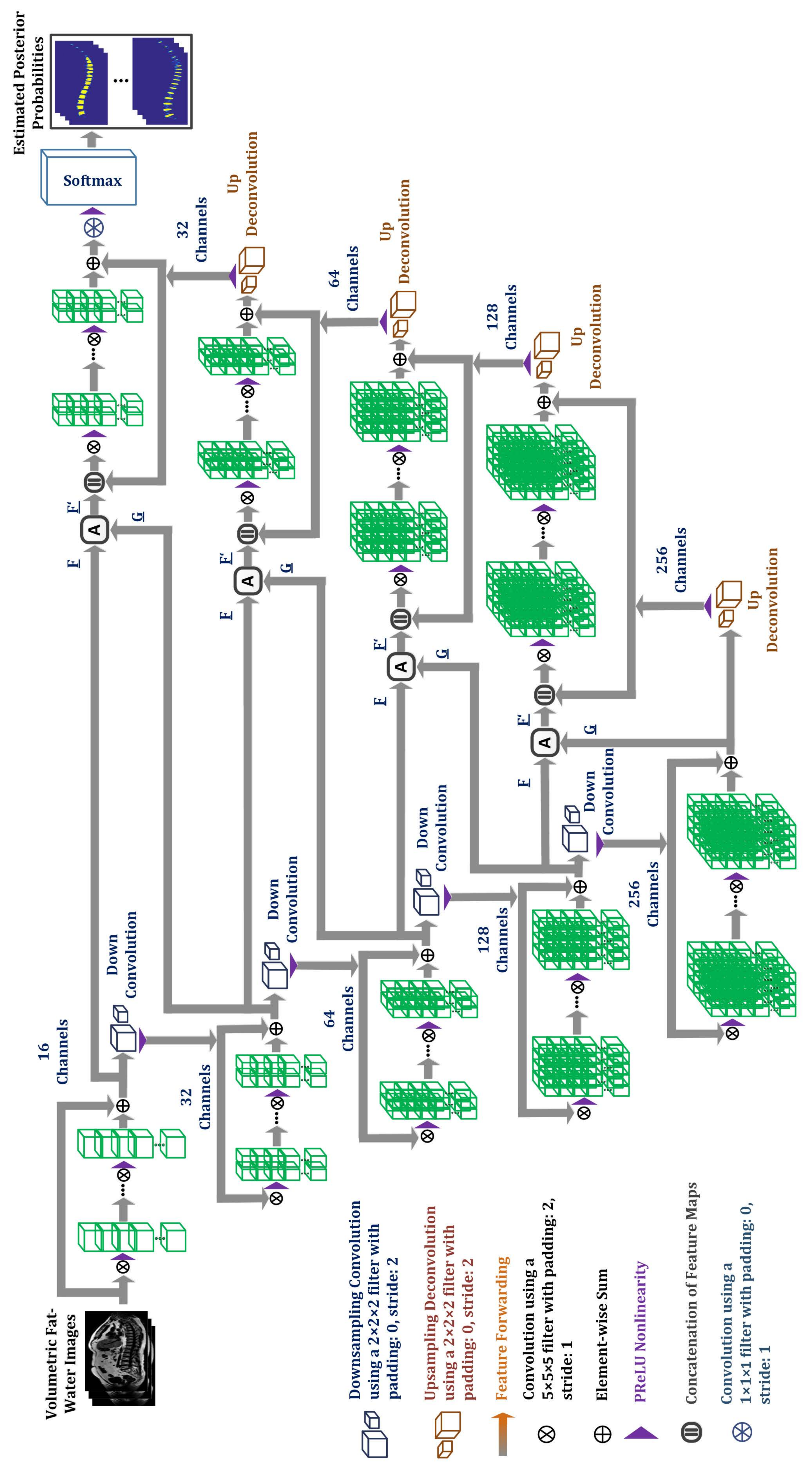}
\caption{Schematic of the 3D fully convolutional and residual V-net with the grid-based gating module in each of its feature forwarding paths.}
\label{fig:VnetAttention}
\end{center}
\end{figure}

In \cite{Oktay2018,Zuo2021,Sijia2022}, to reduce the number of the parameters and the computational complexity of the attention gates, each filter was a convolutional layer of $0$ padding and $1\times 1\times 1$ kernel size, i.e. without any spatial support. To downsample the input feature maps of each attention gate to the resolution of its gating maps, the convolutional filters of the feature maps had a stride of $2$ in each dimension. Moreover, each attention gate handled a binary classification and thus computed a common attention map for all the feature maps at its input. To this end, the downsampling convolutional filters of the feature maps linearly transformed them to an intermediate number of feature maps denoted by C'. Also, the convolutional filters of the gating maps linearly transformed them to C' intermediate maps. The intermediate feature/gating maps were to be more semantically discriminative than the original feature/gating maps in localizing the target structures. Thus, the number C' was a resolution-specific hyperparameter and needed to be optimized for each attention gate separately. Then, according to an \textbf{additive attention} mechanism, the intermediate downsampled feature maps got added to the intermediate gating maps and then passed through a nonlinear rectified linear unit (ReLU), a $1\times 1\times 1$ convolutional layer of $0$ padding and a stride of $1$, and a nonlinear Sigmoid layer to form the attention map for all the input feature maps. This attention map had a lower resolution than the input feature maps and thus was upsampled by a grid-based trilinear interpolation to the same resolution as the input feature maps. In comparison to a multiplicative attention, the additive attention was more computationally demanding but more effective in enhancing the classification accuracy.

To handle a multiclass classification over $n_{\mathrm{clas}}=|\mathbb{L}|$ classes, we modified the aforementioned gating module by replacing the nonlinear Sigmoid function with a nonlinear Softmax function. Also, after the ReLU operation, the $1\times 1\times 1$ convolutional layer did not map the outputs of the ReLU to one channel rather to the number of feature maps at the input of the gating module. That is, instead of computing one common attention map for all the input feature maps, we computed an attention map for each feature map separately and independently from other feature maps. Furthermore, to simplify the network's optimization we eliminated the resolution-specific hyperparameter C' defining the number of the intermediate feature/gating maps. To this end, the $1\times1\times1$ convolutional layer directly applied to the input feature maps transferred them to the number of channels already existing in the input gating maps. This in turn eliminated the $1\times1\times1$ convolutional layer directly applied to the input gating maps and thus further simplified the architecture of the gating module. \autoref{fig:AttentionGate} compares the original gating module with our proposed one and \autoref{fig:VnetAttention} shows the V-net architecture with such a gating module in each of its feature forwarding paths.

To reduce the overfitting of the baseline architectures to the seen (training) samples and thereby improve the generalization (predictive performance on unseen samples), we applied Dropout to every perceptron (node) of these architectures. This technique had a common root with a Bayesian neural network which, as described in \autoref{ssec:optNeuralNetClass}, was an ensemble of plain neural networks. In the training (optimization) phase, the Dropout dropped some of the perceptrons (nodes) of the network by vanishing their incoming and outgoing weights. The keep (retention) probability of each perceptron (node) was the occurrence probability of a Bernoulli distributed random variable. This probability was handled like a tunable hyperparameter indicating the confidence (inverse of the variance) of the node's estimations. We considered a common retention probability for all the perceptrons (nodes) of each encoder/decoder stage of the baseline architectures. For the $s^{\mathrm{th}}$ encoder/decoder stage, this probability was denoted by $p_s\in[0,1]$. In the test phase, all the perceptrons (nodes) of the network were kept. However, the outgoing weights of each node got multiplied by its retention probability optimized during the hyperparameter optimization. The Dropout was shown to be superior to other regularization techniques such as the weight decay which penalized the weights of large $l_2$ norms. This superiority come at the cost of a higher number of iterations for convergence of the optimizations \cite{Srivastava2014,Gal2015,Jospin2022}.

\section{Outline of Contributions}
\label{sec:outContNeuralNet}
All the \textbf{metric-based} losses introduced in \autoref{ssec:CommObjsBotlncks} were independent of the class-sample distribution of the training samples and could thus enhance the generalization (predictive performance on unseen samples) of a neural network trained (optimized) with them. However, the metrics involved in those losses were binary classification metrics. This implied to decompose a multiclass classification into a series of one-vs-all classifications and then form its overall loss from an average of the one-vs-all losses. This was observable in the definition of the DICE loss in \eqref{eq:DICEloss} and the Lov\'{a}sz-Softmax loss in \eqref{eq:lovaszSoftMult}.

The averaging across the classes could naturally lead to a bias towards the dominant classes, i.e. classes of more samples. This bias could not be mitigated by a weighting mechanism such as the ones incorporated in the distribution-based losses introduced in page~\pageref{eq:WCE} and page~\pageref{eq:wghtFocLoss}. The reason was that such a weighting could diminish the false positive mispredictions on dominant classes and could thus mislead the optimization. Moreover, if a class was absent in both the reference labels and the predicted labels, then $\text{DICE}=\text{JI}=1$ and $\text{JD}=0$.

All the \textbf{distribution-based} losses introduced in \autoref{ssec:CommObjsBotlncks} were based on the cross entropy and had a common root with the variational free energy (VFE) of a retrospective active inference. These losses fitted the network's model to the class-sample distribution of the training samples and could thus compromise the network's generalization when the distribution of unseen (validation or test) samples differed from the distribution of the seen (training) samples. However, as described in page~\pageref{eq:WCE} and page~\pageref{eq:wghtFocLoss}, these losses could reduce the classification biases towards the dominant classes by weighting each class's term with regard to its number of samples or importance. In spite of this capability, there existed no optimal weighting which could be incorporated into the cross entropy-based losses to make them equivalent to any of the metric-based losses. Thus, to benefit from the advantages of the cross entropy-based and the metric-based losses while mitigating their drawbacks, a combination of them was used. Alternatively, to reduce the overfitting and thus to improve the generalization of the cross entropy-based losses, additional co-training with augmented training samples got conducted. Also, to reduce the classification biases towards the dominant classes, the false positive mispredictions of the network trained with the metric-based losses got post-corrected by using morphological operations \cite{Isensee2018,Bertels2019,Jadon2020,Chen2022}.

Despite of some improves, all the aforementioned schemes imposed extra overheads to the training or predictions of the neural networks. In addition, the augmentation of the training samples obtained from images was mostly done on the fly by applying gamma (luminance) modifications, mirroring, random scaling, random rotation, and random elastic deformation\footnote{The elastic deformations were obtained from a B-spline interpolation over a grid of control points on a dense deformation field.} to the original images. These techniques could not be easily applied to medical images where pathological alterations should be differentiated from the augmentations. Moreover, none of the aforementioned schemes could completely mitigate the overfitting of a large network to a limited number of the training samples or the classification biases towards the dominant classes. Furthermore, none of the described losses could \textbf{incorporate priors} or \textbf{handle errors or uncertainties in the reference labels of the training samples} \cite{Lo2021}.
Errors in the reference labels of the training samples could arise from human errors in the manual annotations of the training samples and images or the errors induced by noise and artifacts. Uncertainties and ambiguities in the reference labels of the training samples could stem from similar features and textures of different classes. These similarities not only confused the manual annotators but also the neural network relying on those features and textures for learning boundaries between different classes.

To mitigate the aforementioned bottlenecks, we proposed
\begin{enumerate}[label={(\arabic*)},leftmargin=*]
\item a novel algorithm, based on the generalized (multinomial) Kelly criterion for optimal betting, to recompute the reference labels of the training samples by using their priors and the currently estimated classification posteriors on the network;
\item a novel objective function, based on the expected free energy (EFE) of a prospective active inference, with the capability of
\begin{itemize}
\item incorporating prior probabilities of the training samples to focus the attention of the neural network on important but minority foreground classes and thereby reshape the effectively seen distribution for a reduction of the class-sample unbalancedness, the overfitting, and the classification biases towards the dominant classes;
\item representing the \textit{precision} and \textit{recall} metrics by its terms to enhance the robustness of the network's optimization against the class-sample unbalancedness;
\end{itemize}
\item a process to integrate the proposed algorithm and the proposed objective function into a mini-batch-based gradient descent optimizer with backpropagation.
\end{enumerate}

The proposed algorithm for recomputing the reference labels was listed in Algorithm \ref{Lopt}. This algorithm calculated a \textbf{set of candidate labels} for each training sample from its prior and currently estimated posterior probabilities on the network. This algorithm resulted from our reformulation of the generalized (multinomial) Kelly criterion for optimal betting on multiple horses in a horse race. This reformulation cast the generalized Kelly criterion into a multiclass classification problem by interpreting each training sample as a bettor, each class as a horse, and each iteration of the network's optimization as a horse race. Then, the classification prior of the training sample with regard to each class become the win probability of the corresponding horse. The classification posterior currently estimated by the network for the training sample with regard to the same class become the belief probability of the corresponding horse. The proposed sets of candidate labels got then plugged into the proposed objective function to form the current loss for an update (optimization) of the network's parameters in the current iteration. Thus, instead of a reference label, a set of candidate labels got considered for each training sample in each iteration.

This consideration allowed to mitigate the aforementioned uncertainties and ambiguities in the labels generated from manual annotations in the presence of noise, artifacts, and similar features or textures of different classes. In other words, the sets of candidate labels could handle possible overlaps between different classes and thus enhanced the reliability and the flexibility of the neural network's optimization. More specifically, these sets could help a gradient descent optimizer to escape from local optimums caused by the original reference labels. Moreover, if the reference labels of some training samples were missing, then their candidate labels could still be computed from their priors and posteriors. This \textbf{semi-supervised optimization} was of particular importance in the applications where the manual annotations of the reference labels were costly and cumbersome.

Our proposed Algorithm \ref{Lopt} for finding the candidate labels aimed to minimize the objective function of the generalized Kelly criterion. This minimized function was given by \eqref{eq:minKellyObj} and was indeed the \textbf{expected~complexity} term of the EFE of a prospective active inference. That is, the objective function of the generalized Kelly criterion was a tight upper bound of the expected~complexity of the EFE. The EFE was given by \eqref{eq:objFuncEFE} and was composed of an \textbf{expected~complexity} term plus an \textbf{uncertainty} term. As described in \autoref{ssec:activeInference}, the minimization of the expected~complexity was equivalent to the maximization of the reward. The reward maximization was also a goal of the Kelly criterion and could thus be partially fulfilled by finding the candidate labels through the proposed Algorithm \ref{Lopt}.

More specifically, from the prior (win) and the posterior (belief) probabilities of each training sample (bettor), the generalized Kelly criterion computed optimal allocation fractions of the bettor's asset for betting on the candidate classes (horses)\footnote{The allocation fractions for noncandidate classes (horses) were zero.}. These allocation fractions maximized the geometric average of the growth rate of the bettor's asset or the \textbf{reward}. To further maximize the reward, the expected~complexity of the EFE should be minimized further. This was doable by having enough information or maximizing the information gain, i.e. minimizing the \textbf{uncertainty} of the EFE. Accordingly, to optimize a discriminative neural network classifier, we proposed a novel objective function based on the EFE of a prospective active inference. This function was given by \eqref{eq:propObjFunc} and was reversible and differentiable with respect to the outputs of every layer of the neural network. Thus, as described in \autoref{ssec:optNeuralNetClass}, it could be minimized by a gradient descent optimizer with backpropagation.

As explained in \autoref{ssec:CommObjsBotlncks}, all the cross entropy-based losses were \textbf{distribution-based} and stemmed from the VFE given by \eqref{eq:objFuncVFE} for a retrospective active inference. The VFE was \textbf{complexity} minus \textbf{accuracy}. The complexity reflected the overfitting of the neural network's model to the distribution of seen (training) samples and thus the variance of the predictions on unseen (validation or test) samples. The accuracy was inversely proportional to the bias (difference) of the predictions from their true values. Thus, the minimization of the VFE implied to minimize the complexity or the overfitting while maximizing the classification accuracy by minimizing the classification bias. This way, the VFE and the cross entropy-based losses addressed the bias-variance tradeoff of the classification problems without considering the unbalancedness of the class-sample distribution of the seen samples.

In contrast, the EFE given by \eqref{eq:objFuncEFE} for a prospective active inference and thus our proposed objective function in \eqref{eq:propObjFunc} addressed the unbalancedness of the class-sample distribution of the seen (training) samples by representing the \textit{precision} and \textit{recall} metrics in their terms. The \textit{precision} and the \textit{recall} metrics were independent of the correct classification of unimportant majority samples (designated by true negatives) and instead focused on the correct classification of important minority samples (designated by true positives). This made them less sensitive than the other metrics to the imbalanced class-sample distributions and the classification biases towards the dominant classes.

As mentioned earlier, the minimization of the EFE or our proposed objective function implied to minimize the \textbf{expected~complexity} and the \textbf{uncertainty}. The minimization of the expected complexity implied to maximize the reward and the reward was equivalent to the \textit{recall} (completeness or diversity). The minimization of the uncertainty implied to maximize the information gain or the \textit{precision} (exactness or confidence). This way, the EFE and our proposed objective function aimed to maximize the \textit{precision} and the \textit{recall} metrics. This allowed them to handle an imbalanced class-sample distribution while still being \textbf{distribution-based} \cite{Feldman2010,Tatbul2018,Smith2022}.

Moreover, our proposed objective function could incorporate the prior probabilities of the training samples directly and indirectly. The indirect incorporation was through using the candidate classification labels computed from the priors and the posteriors of the training samples by the proposed Algorithm \ref{Lopt}. This incorporation resulted in a grouping of the terms of the proposed objective function with regards to the candidate and noncandidate labels. More specifically, the priors or the posteriors of the noncandidate labels got summed together to form a collective prior or posterior for the noncandidate classes. This way, the noncandidate classes formed a collective class together and the neural network got enforced to find the boundary between each candidate class and the collective class of the noncandidates. In comparison to computing the boundaries between each pair of the classes, this grouping reduced the effective number of the classes and the boundaries needed to be computed. This in turn reduced the network's complexity and its overfitting to the seen (training) distribution and could thus enhance its generalization (predictive performance on unseen samples).

The direct incorporation of the prior probabilities of the training samples into the objective function of the network's optimization could focus the attention of the neural network on important but minority foreground classes. This could reshape the distribution effectively seen by the network during its optimization and could thereby reduce the class-sample unbalancedness, the overfitting, and the classification biases towards the dominant classes \cite{Maier2019}. Similar effects could result from the architecture-based attention mechanisms described in \autoref{ssec:baselineArchi}. That is, if no prior probabilities were provided, then \textbf{stronger posteriors} resulted from an \textbf{architecture-based attention mechanism} should help.
In the baseline architecture described in \autoref{ssec:baselineArchi}, an attention gate could be incorporated into each feature forwarding path between an encoder stage and its corresponding decoder stage. Without such a gate, the feature forwarding path was a plain skip connection.

Our proposed algorithm for finding the candidate labels and our proposed objective function for optimizing a discriminative neural network classifier got integrated into a mini-batch-based gradient descent optimizer with backpropagation by using the process proposed in \autoref{sec:PropOptProc}. This process got evaluated against a similar process incorporating a representative of the cross entropy-based losses or a representative of the metric-based losses introduced in \autoref{ssec:CommObjsBotlncks}. The representative of the cross entropy-based losses was the weighted focal loss. This loss comprised of a modulating factor and a weighting mechanism to alleviate classification biases towards the dominant classes of the training samples. The representative of the metric-based losses was the Lov\'{a}sz-Softmax loss. Besides being smooth and differentiable, to the best of our knowledge, this loss was the only convex loss among the metric-based losses.

Accordingly, the evaluated losses were
\begin{enumerate}[label={(\arabic*)},leftmargin=*]
\item the proposed objective function given by \eqref{eq:propObjFunc}
\item the weighted focal loss given by \eqref{eq:wghtFocLoss}
\item the Lov\'{a}sz-Softmax loss given by \eqref{eq:lovaszSoftMult}.
\end{enumerate}
These evaluations were on an end-to-end optimization of the baseline architecture described in \autoref{ssec:baselineArchi}. For each case, the baseline architecture was once used without attention gates as depicted in \autoref{fig:Vnet} and once used with the attention gates as depicted in \autoref{fig:VnetAttention}. Also, for (2) and (3) each training sample was accompanied by its reference (ground truth) label to fulfill the supervised nature of these objective functions. However, our proposed algorithm for finding the candidate labels and our proposed objective function got evaluated according to a fully supervised, a semi-supervised, and an unsupervised approach. These resulted in the training samples being
\begin{enumerate}[label={(\arabic*)},leftmargin=*]
\item accompanied by their reference labels and their priors~$\rightarrow$~fully supervised
\item only accompanied by their reference labels~$\rightarrow$~semi-supervised
\item only accompanied by their priors~$\rightarrow$~semi-supervised
\item accompanied by neither their reference labels nor their priors~$\rightarrow$~unsupervised.
\end{enumerate}
The unsupervised case only relied on the posteriors estimated by the neural network during its optimization and could thus be considered as a self-supervised case as well.

For the cases with the priors, the prior probabilities of the training samples could be computed by a multiatlas registration. If no prior probabilities were provided at the time of optimization (training), then uniform priors got assumed. If the reference (ground truth) labels of the training samples $\mathbb{T}_{\mathrm{train}}$ were provided at the time of optimization (training), then for each sample $v_{b,j}\in\mathbb{T}_b\subseteq\mathbb{T}_{\mathrm{train}}$ the vectorized reference label $\mathbf{l}_{b,j}$ was the one-hot-encoding of its reference label $l_{b,j}\in\mathbb{L}$ and was given by \eqref{eq:refLabelOneHot}. If the reference labels of the training samples $\mathbb{T}_{\mathrm{train}}$ were not provided at the time of optimization, then for each sample $v_{b,j}\in\mathbb{T}_b\subseteq\mathbb{T}_{\mathrm{train}}$ the vector $\mathbf{l}_{b,j}$ was uniform and given by \eqref{eq:refLabelUniform}.

For each evaluation case, the main parameters and the hyperparameters of the baseline architecture got trained (optimized) to automatically segment $n_{\mathrm{clas}}=|\mathbb{L}|=8$ classes of vertebral bodies (VBs), intervertebral disks (IVDs), psoas major (PM) and quadratus lumborum (QL) muscles, epicardial adipose tissues (EpAT), pericardial adipose tissues (PeAT), cardiac perivascular adipose tissues (PvAT), and background on each volumetric fat-water image. To this end, the volumetric fat-water images got divided into a training and a test set. The training set formed the samples set $\mathbb{T}_{\mathrm{train}}$ and got used to optimize the main parameters and the hyperparameters of the baseline architecture by each method. The test set formed the samples set $\mathbb{T}_{\mathrm{test}}$ and got used to evaluate the classification performance of the baseline architecture after being fully optimized by each method. The training set was composed of samples accompanied by their reference labels and priors. The test set was composed of samples accompanied by their reference labels. The reference labels of the test samples were not fed to the neural network. They were rather compared against the corresponding labels predicted by the network to evaluate the classification performance of the network. The predicted label of each sample was the index of its maximum classification posterior estimated by the network.

Finally, our proposed optimization process was based on the generalized Kelly criterion for optimal betting and a prospective active inference. It addressed optimization of discriminative neural network classifiers with a feed-forward architecture. Active inference-based optimizations could foster building highly flexible and generalizable generative models with and without memory. An example of a model with the memory was the one which could explain a partially observable Markov decision process. This model could be implemented by a recurrent or a long short-term memory network \cite{Friston2019,Smith2022,Carr2022}. Accordingly, our proposed optimization process could be easily extended to generative or recurrent neural networks such as the networks in \cite{Alom2019,Zuo2021,Mehreen2022}.

\section{Application of the Kelly Criterion to Classification}
\label{sec:GenKelly}
The generalized (multinomial) Kelly criterion proposed optimal allocation fractions of a bettor's asset in betting on multiple horses in a horse race. Each horse had a win and a belief probability. The win probability was the chance of the horse to win the race. The belief probability was the collective belief of other bettors about the chance of the horse to win the race. Thus, for a specific bettor, an optimum betting strategy was to invest as much as possible on a horse of maximum win probability and minimum belief probability (minimum number of other bettors investing on it). This was based on the assumption that all the bettors followed the same strategy and the gain of a horse win got divided between all the bettors who have invested on it. Therefore, the lesser the belief probability was, the higher the paid gain to the investing bettor would be \cite{Kelly1956,Smoczynski2010}.

To optimize a discriminative neural network classifier in a multiclass classification over $n_{\mathrm{clas}}=|\mathbb{L}|$ classes by using the generalized Kelly criterion, we assumed
\begin{itemize}[leftmargin=*]
\item every training sample $v_{b,j}\in\mathbb{T}_b\subseteq\mathbb{T}_{\mathrm{train}}$ to be a bettor
\item every class $c\in\mathbb{L}$ to be a horse
\item every iteration $i\in\{1,\cdots,n_{\mathrm{it}}\}$ of the optimization to be a round of horse race with its gambling competitions among the bettors (training samples)
\item the win probability of each horse (class) $c\in\mathbb{L}$ for each bettor (training sample) $v_{b,j}\in\mathbb{T}_b\subseteq\mathbb{T}_{\mathrm{train}}$ to be the prior probability $a_{b,j,c}\in(0,1)$ estimated by another classifier\footnote{If no prior probabilities were provided then uniform priors got assumed.}
\item the belief probability of each class $c\in\mathbb{L}$ for each sample $v_{b,j}\in\mathbb{T}_b\subseteq\mathbb{T}_{\mathrm{train}}$ to be the classification posterior $\hat{p}_{b,j,c}^{(i)}\in(0,1)$ estimated by the network in the current iteration $i$.
\end{itemize}
It should be noted that in the betting, the win probabilities of the horses were shared across the bettors, but, in the classification, each sample had its own win probability for each class. Moreover, the interpretation of the estimated posteriors of the network as the belief probabilities might look counterintuitive because each sample (bettor) had no \textit{other} samples (bettors) to compete with. Thus the overall belief about a class (horse) could not be collected from other samples (bettors). Moreover, it was more tempting to select a class (invest on a horse) of maximum belief probability as this probability could be an indicator of the chance of the class (horse) to win. Our definition of the win probability and our counterintuitive definition of the belief probability could be explained under an \textbf{attention mechanism}.

On one hand, the selection of the classes (horses) of maximum win probability encouraged the network to focus on classes of confident (high) prior probabilities. In an image segmentation task conducted in a spatial domain, this implied to focus on important (relevant) regions highlighted by high prior probabilities in the image. On the other hand, the selection of the classes (horses) of minimum belief probability encouraged the network to focus on inconfident (low) posteriors and thus to improve its classification by tackling difficult examples.

In each iteration (race) $i$, for each training sample (bettor) $v_{b,j}\in\mathbb{T}_b\subseteq\mathbb{T}_{\mathrm{train}}$, the Kelly criterion proposed allocation fractions $\hat{\mathbf{g}}_{b,j}^{(i)}={\big[\hat{g}_{b,j,c}^{(i)}\in[0,1]\big]}_{c\in\mathbb{L}}$ of its asset for betting on $n_{\mathrm{clas}}=|\mathbb{L}|$ classes (horses). If in the iteration (race) $i$ the class (horse) $c\in\mathbb{L}$ won, then the asset of $v_{b,j}\in\mathbb{T}_b\subseteq\mathbb{T}_{\mathrm{train}}$ would be multiplied by $\Big[1-\sum_{k\in\mathbb{L}}\hat{g}_{b,j,k}^{(i)}+\frac{\hat{g}_{b,j,c}^{(i)}}{\hat{p}_{b,j,c}^{(i)}}\Big]^{-1}$. We assumed that the outcomes of the iterations (horse races) were independent identically distributed (i.i.d.) random variables. Thus, after $i$ iterations, the geometric average of the growth rate of the asset of $v_{b,j}\in\mathbb{T}_b\subseteq\mathbb{T}_{\mathrm{train}}$ with $n_c^{(i)}\in[0,i]$ number of wins for each class $c\in\mathbb{L}$ become
\begin{equation}
\label{eq:geomAvgGrthRate}
\eta_{b,j}^{(i)}=\prod_{c\in\mathbb{L}}\bigg[1-\sum_{k\in\mathbb{L}}\hat{g}_{b,j,k}^{(i)}+\frac{\hat{g}_{b,j,c}^{(i)}}{\hat{p}_{b,j,c}^{(i)}}\bigg]^{-n_c^{(i)}/i}~~~~~~~i=\sum_{c\in\mathbb{L}}n_c^{(i)}.
\end{equation}

By taking the $\mathrm{ln}(\cdot)$ of both sides of \eqref{eq:geomAvgGrthRate}, one obtained
\begin{equation}
\label{eq:geomAvgGrthRateLn}
\begin{gathered}
\mathrm{ln}(\eta_{b,j}^{(i)})=\sum_{c\in\mathbb{L}}{\frac{-n_c^{(i)}}{i}\cdot\mathrm{ln}\bigg[1-\sum_{k\in\mathbb{L}}\hat{g}_{b,j,k}^{(i)}+\frac{\hat{g}_{b,j,c}^{(i)}}{\hat{p}_{b,j,c}^{(i)}}\bigg]}\\
\lim_{i\to\infty}\frac{n_c^{(i)}}{i}=a_{b,j,c}\implies\lim_{i\to\infty}\mathrm{ln}(\eta_{b,j}^{(i)})=\sum_{c\in\mathbb{L}}{-a_{b,j,c}\cdot\mathrm{ln}\bigg[1-\sum_{k\in\mathbb{L}}\hat{g}_{b,j,k}^{(i)}+\frac{\hat{g}_{b,j,c}^{(i)}}{\hat{p}_{b,j,c}^{(i)}}\bigg]}.
\end{gathered}
\end{equation}
If the allocation fractions $\mathbf{g}_{b,j}^{(i)}={\big[g_{b,j,c}^{(i)}\in[0,1]\big]}_{c\in\mathbb{L}}$ proposed by the Kelly criterion for each sample (bettor) $v_{b,j}\in\mathbb{T}_b\subseteq\mathbb{T}_{\mathrm{train}}$ were \textbf{asymptotically optimum} over a long run $(i\to\infty)$, then they maximized the geometric average in \eqref{eq:geomAvgGrthRate}. Due to the monotonic increase of the $\mathrm{ln}(\cdot)$ function, the maximization of \eqref{eq:geomAvgGrthRate} was equivalent to the maximization of \eqref{eq:geomAvgGrthRateLn}. This way, the asymptotically optimum allocation fractions were the maximizers of the averaged logarithms of the growth rate in \eqref{eq:geomAvgGrthRateLn}. That is, $\mathbf{g}_{b,j}^{(i)}=\argmax_{\hat{\mathbf{g}}_{b,j}^{(i)}}~\Big[\mathrm{ln}(\eta_{b,j}^{(i)})\Big]$ or
\begin{equation}
\label{eq:optAllocFrac}
\mathbf{g}_{b,j}^{(i)}=\argmin_{\hat{\mathbf{g}}_{b,j}^{(i)}}~\Big[-\mathrm{ln}(\eta_{b,j}^{(i)})\Big]=\argmin_{\hat{\mathbf{g}}_{b,j}^{(i)}}~\Bigg[\sum_{c\in\mathbb{L}}{a_{b,j,c}\cdot\mathrm{ln}\bigg[1-\sum_{k\in\mathbb{L}}\hat{g}_{b,j,k}^{(i)}+\frac{\hat{g}_{b,j,c}^{(i)}}{\hat{p}_{b,j,c}^{(i)}}\bigg]}\Bigg].
\end{equation}

As detailed in \cite{Smoczynski2010}, $\hat{\mathbf{g}}_{b,j}^{(i)}={\big[\hat{g}_{b,j,c}^{(i)}\big]}_{c\in\mathbb{L}}\in[0,1]^{n_{\mathrm{clas}}=|\mathbb{L}|}$ formed a convex set
\begin{equation}
\label{eq:convSetKelly}
\mathbb{G}_{b,j}^{(i)}=\Bigg\{\hat{\mathbf{g}}_{b,j}^{(i)}\in[0,1]^{n_{\mathrm{clas}}=|\mathbb{L}|}~\bigg|~\bigg[1-\sum_{k\in\mathbb{L}}\hat{g}_{b,j,k}^{(i)}+\frac{\hat{g}_{b,j,c}^{(i)}}{\hat{p}_{b,j,c}^{(i)}}\bigg]>0\Bigg\}\subseteq[0,1]^{n_{\mathrm{clas}}=|\mathbb{L}|}
\end{equation}
which was an intersection of half spaces. Each half space was a side of a hyperplane. In addition, in the above optimization, $\big[1-\sum_{k\in\mathbb{L}}\hat{g}_{b,j,k}^{(i)}\big]\in[0,1]\implies\sum_{k\in\mathbb{L}}\hat{g}_{b,j,k}^{(i)}\in[0,1]$. That is, it was allowed to back a horse to win but not to lay a horse to lose. This condition constrained every $\hat{\mathbf{g}}_{b,j}^{(i)}\in\mathbb{G}_{b,j}^{(i)}$ to a stricter convex set given by
\begin{equation}
\label{eq:convSetMax}
{\mathbb{G}'}_{b,j}^{(i)}=\Big\{\hat{\mathbf{g}}_{b,j}^{(i)}\in\mathbb{G}_{b,j}^{(i)}~\Big|~\sum_{k\in\mathbb{L}}\hat{g}_{b,j,k}^{(i)}\leq 1~~\text{and}~~\forall c\in\mathbb{L}:\hat{g}_{c,j}^{(i)}\geq0\Big\}\subseteq\mathbb{G}_{b,j}^{(i)}.
\end{equation}

The definition of $\mathrm{ln}(\eta_{b,j}^{(i)})$ in \eqref{eq:geomAvgGrthRateLn} showed that it was a finite linear combination of strictly concave logarithms with the coefficients being the priors $\mathbf{a}_{b,j}={\big[a_{b,j,c}\in(0,1)\big]}_{c\in\mathbb{L}}$. This way, the $\mathrm{ln}(\eta_{b,j}^{(i)})$ become differentiable, strictly concave downwards, and of a unique maximum on the boundary of every bounded subset of $\mathbb{G}_{b,j}^{(i)}$. Accordingly, to find the maximizers of $\mathrm{ln}(\eta_{b,j}^{(i)})$ or the optimum allocation fractions $\mathbf{g}_{b,j}^{(i)}={\big[g_{b,j,c}^{(i)}\in[0,1]\big]}_{c\in\mathbb{L}}$, it was enough to only explore the boundaries of ${\mathbb{G}'}_{b,j}^{(i)}\subseteq\mathbb{G}_{b,j}^{(i)}$ \cite{Smoczynski2010}. This exploration (maximization) could be done by using the method of Lagrange multipliers and the Karush-Kuhn-Tucker (KKT) theory \cite{Boyd2004}. That is, instead of maximizing $\mathrm{ln}(\eta_{b,j}^{(i)})$, we maximized
\begin{equation}
\label{eq:lagrMultObj}
\gamma_{b,j}^{(i)}=\mathrm{ln}(\eta_{b,j}^{(i)})+\Big[\sum_{k\in\mathbb{L}}\lambda_{b,j,k}^{(i)}\cdot\hat{g}_{b,j,k}^{(i)}\Big]+\lambda_{b,j,0}^{(i)}\cdot\Big[1-\sum_{k\in\mathbb{L}}\hat{g}_{b,j,k}^{(i)}\Big]
\end{equation}
with $\big\{\lambda_{b,j,k}^{(i)}\in\mathbb{R}_{\geq 0}\big\}_{k=0}^{|\mathbb{L}|}$ being the Lagrange multipliers.

The KKT theory stated that every constrained maximizer of $\mathrm{ln}(\eta_{b,j}^{(i)})$ was an unconstrained maximizer of $\gamma_{b,j}^{(i)}$. The unconstrained maximization of $\gamma_{b,j}^{(i)}$ was done through vanishing its gradient (derivatives) with respect to $\hat{\mathbf{g}}_{b,j}^{(i)}={\big[\hat{g}_{b,j,c}^{(i)}\in[0,1]\big]}_{c\in\mathbb{L}}$. That is,
\begin{equation}
\label{eq:gradLagMultObj}
\frac{\partial\gamma_{b,j}^{(i)}}{\partial\hat{g}_{b,j,c}^{(i)}}=\frac{-a_{b,j,c}+a_{b,j,c}/\hat{p}_{b,j,c}^{(i)}}{1-\sum_{k\in\mathbb{L}}\hat{g}_{b,j,k}^{(i)}+\hat{g}_{b,j,c}^{(i)}/\hat{p}_{b,j,c}^{(i)}}+\lambda_{b,j,c}^{(i)}-\lambda_{b,j,0}^{(i)}=0.
\end{equation}

This resulted in the following KKT optimality constraints:
\begin{equation}
\label{eq:KKToptConds}
\begin{split}
\text{if}~~~\lambda_{b,j,c}^{(i)}\cdot\hat{g}_{b,j,c}^{(i)}=0&\implies\lambda_{b,j,c}^{(i)}=0~~~\text{if}~~~\hat{g}_{b,j,c}^{(i)}>0\\
\text{if}~~~\lambda_{b,j,0}^{(i)}\cdot\Big[1-\sum_{k\in\mathbb{L}}\hat{g}_{b,j,k}^{(i)}\Big]=0&\implies\lambda_{b,j,0}^{(i)}=0~~~\text{if}~~~\sum_{k\in\mathbb{L}}\hat{g}_{b,j,k}^{(i)}<1.
\end{split}
\end{equation}

The allocation fractions $\hat{\mathbf{g}}_{b,j}^{(i)}={\big[\hat{g}_{b,j,c}^{(i)}\in[0,1]\big]}_{c\in\mathbb{L}}$ and the Lagrange multipliers $\big\{\lambda_{b,j,k}^{(i)}\in\mathbb{R}_{\geq 0}\big\}_{k=0}^{|\mathbb{L}|}$ should fulfill \eqref{eq:KKToptConds} on the convex set ${\mathbb{G}'}_{b,j}^{(i)}\subseteq\mathbb{G}_{b,j}^{(i)}$. According to \cite{Smoczynski2010}, the maximum of $\mathrm{ln}(\eta_{b,j}^{(i)})$ under $\sum_{k\in\mathbb{L}}\hat{g}_{b,j,k}^{(i)}=1$ was less than its maximum under $\sum_{k\in\mathbb{L}}\hat{g}_{b,j,k}^{(i)}<1$. Thus, in \eqref{eq:convSetMax}, we replaced $\sum_{k\in\mathbb{L}}\hat{g}_{b,j,k}^{(i)}\leq 1$ with $\sum_{k\in\mathbb{L}}\hat{g}_{b,j,k}^{(i)}<1$ and obtained $\lambda_{b,j,0}^{(i)}=0$ from \eqref{eq:KKToptConds}. For each sample (bettor) $v_{b,j}\in\mathbb{T}_b\subseteq\mathbb{T}_{\mathrm{train}}$, the classes (horses) whose allocation fractions were nonzero were deemed to be \textbf{candidate} and formed the set $\mathbb{L}_{b,j}^{(i)}$ with
\begin{equation}
\label{eq:candidSet}
\begin{split}
\forall c\in\mathbb{L}_{b,j}^{(i)}\subseteq\mathbb{L}:&~~~\hat{g}_{b,j,c}^{(i)}>0~~~\text{and}~~~\lambda_{b,j,c}^{(i)}=0\\\forall c\in\mathbb{L}-\mathbb{L}_{b,j}^{(i)}:&~~~\hat{g}_{b,j,c}^{(i)}=0~~~\text{and}~~~\lambda_{b,j,c}^{(i)}\geq 0.
\end{split}
\end{equation}

Then, solving \eqref{eq:gradLagMultObj} under the above conditions gave
\begin{equation}
\label{eq:optFracClosed}
\forall c\in\mathbb{L}_{b,j}^{(i)}\subseteq\mathbb{L}:~g_{b,j,c}^{(i)}=a_{b,j,c}-\hat{p}_{b,j,c}^{(i)}\cdot\frac{\sum_{k\in\mathbb{L}-\mathbb{L}_{b,j}^{(i)}}a_{b,j,k}}{\sum_{k\in\mathbb{L}-\mathbb{L}_{b,j}^{(i)}}\hat{p}_{b,j,k}^{(i)}}
\vspace{-3mm}
\end{equation}
\begin{subequations}
\label{eq:unspentAsset}
\begin{align*}
\implies s_{b,j}^{(i)}&=1-\sum_{c\in\mathbb{L}}g_{b,j,c}^{(i)}=1-\sum_{c\in\mathbb{L}_{b,j}^{(i)}}g_{b,j,c}^{(i)}=\overbrace{1-\sum_{c\in\mathbb{L}_{b,j}^{(i)}}a_{b,j,c}}^{\sum_{k\in\mathbb{L}-\mathbb{L}_{b,j}^{(i)}}a_{b,j,k}}+\frac{\sum_{k\in\mathbb{L}-\mathbb{L}_{b,j}^{(i)}}a_{b,j,k}}{\sum_{k\in\mathbb{L}-\mathbb{L}_{b,j}^{(i)}}\hat{p}_{b,j,k}^{(i)}}\cdot\sum_{c\in\mathbb{L}_{b,j}^{(i)}}\hat{p}_{b,j,c}^{(i)}\\
&=\sum_{k\in\mathbb{L}-\mathbb{L}_{b,j}^{(i)}}a_{b,j,k}\cdot\bigg[1+\frac{\sum_{c\in\mathbb{L}_{b,j}^{(i)}}\hat{p}_{b,j,c}^{(i)}}{\sum_{k\in\mathbb{L}-\mathbb{L}_{b,j}^{(i)}}\hat{p}_{b,j,k}^{(i)}}\bigg]=\frac{\sum_{k\in\mathbb{L}-\mathbb{L}_{b,j}^{(i)}}a_{b,j,k}}{\sum_{k\in\mathbb{L}-\mathbb{L}_{b,j}^{(i)}}\hat{p}_{b,j,k}^{(i)}}\tag{\ref{eq:unspentAsset}}
\end{align*}
\end{subequations}
\begin{equation}
\label{eq:lnArgObj}
\resizebox{0.9\textwidth}{!}{%
$\forall c\in\mathbb{L}_{b,j}^{(i)}\subseteq\mathbb{L}:~s_{b,j}^{(i)}+\frac{g_{b,j,c}^{(i)}}{\hat{p}_{b,j,c}^{(i)}}=\frac{\sum_{k\in\mathbb{L}-\mathbb{L}_{b,j}^{(i)}}a_{b,j,k}}{\sum_{k\in\mathbb{L}-\mathbb{L}_{b,j}^{(i)}}\hat{p}_{b,j,k}^{(i)}}+\frac{a_{b,j,c}}{\hat{p}_{b,j,c}^{(i)}}-\frac{\sum_{k\in\mathbb{L}-\mathbb{L}_{b,j}^{(i)}}a_{b,j,k}}{\sum_{k\in\mathbb{L}-\mathbb{L}_{b,j}^{(i)}}\hat{p}_{b,j,k}^{(i)}}=\frac{a_{b,j,c}}{\hat{p}_{b,j,c}^{(i)}}$%
}
\end{equation}
\begin{equation}
\label{eq:ineqAsset}
\forall c\in\mathbb{L}_{b,j}^{(i)}\subseteq\mathbb{L}~~\text{and}~~\forall l\in\mathbb{L}-\mathbb{L}_{b,j}^{(i)}:~~~\frac{a_{b,j,l}}{\hat{p}_{b,j,l}^{(i)}}\leq s_{b,j}^{(i)}=\frac{\sum_{k\in\mathbb{L}-\mathbb{L}_{b,j}^{(i)}}a_{b,j,k}}{\sum_{k\in\mathbb{L}-\mathbb{L}_{b,j}^{(i)}}\hat{p}_{b,j,k}^{(i)}}<\frac{a_{b,j,c}}{\hat{p}_{b,j,c}^{(i)}}.
\end{equation}

\section{Proposed Objective and Process of Optimization}
\label{sec:PropOptProc}
By using our classification-based formulation of the Kelly criterion in \autoref{sec:GenKelly} we proposed an objective function and a process for optimizing discriminative neural network classifiers. To be generic, we formulated the objective and the process in such a way that they could accommodate a fully supervised, a semi-supervised, or an unsupervised optimization. In the fully supervised optimization, both the reference (ground truth) labels and the prior (win) probabilities of the training samples were provided at the time of optimization (training). In the semi-supervised optimization, either the reference labels or the prior (win) probabilities of the training samples were not provided at the time of optimization (training). In the unsupervised optimization, neither the reference labels nor the prior (win) probabilities of the training samples were provided at the time of optimization (training). If no prior probabilities were provided at the time of optimization (training), then uniform priors got assumed. If the reference (ground truth) labels of the training samples $\mathbb{T}_{\mathrm{train}}$ were provided at the time of optimization (training), then for each sample $v_{b,j}\in\mathbb{T}_b\subseteq\mathbb{T}_{\mathrm{train}}$ the vectorized reference label $\mathbf{l}_{b,j}$ was a one-hot-encoding of its reference (ground truth) label $l_{b,j}\in\mathbb{L}$ and was given by \eqref{eq:refLabelOneHot}. If the reference (ground truth) labels of the training samples $\mathbb{T}_{\mathrm{train}}$ were not provided at the time of optimization (training), then for each sample $v_{b,j}\in\mathbb{T}_b\subseteq\mathbb{T}_{\mathrm{train}}$ the vector $\mathbf{l}_{b,j}$ was uniform and given by \eqref{eq:refLabelUniform}.

\begin{algorithm}[t]
\caption{Determination of the set of candidate classification labels $\mathbb{L}_{b,j}^{(i)}\subseteq\mathbb{L}$}
\label{Lopt}
{\small
\KwIn{Posterior $\hat{\mathbf{p}}_{b,j}^{(i)}={[\hat{p}_{b,j,c}^{(i)}]}_{c\in\mathbb{L}}$ and prior $\mathbf{a}_{b,j}={\big[a_{b,j,c}\in(0,1)\big]}_{c\in\mathbb{L}}$ probabilities}
\KwOut{The set of candidate labels $\mathbb{L}_{b,j}^{(i)}\subseteq\mathbb{L}$ for the sample $v_{b,j}\in\mathbb{T}_b\subseteq\mathbb{T}_{\mathrm{train}}$}
\textbf{Initialization:}~$\mathbb{L}_{b,j}^{(i)}\gets\emptyset,~s_{b,j}^{(i)}\gets 1$\\\vspace{2mm}
\begin{minipage}{0.9\linewidth}
\begin{itemize}[leftmargin=*]
\item Calculate $\Big\{q_{b,j,c}^{(i)}=a_{b,j,c}/\hat{p}_{b,j,c}^{(i)}\Big\}_{c\in\mathbb{L}}$ and sort it in a descending order.
\item Assign the sorted set to $\mathbb{Q}_{b,j}^{(i)}=\Big\{q_{b,j,k}^{(i)}\Big\}_{k=1}^{n_{\mathrm{clas}}=|\mathbb{L}|}$ with $k$ being the sorted index and $\phi_{\mathrm{sort}}:k\rightarrow c$ being a map from the sorted index $k$ to the original index $c$.
\end{itemize}
\end{minipage}\\\vspace{2mm}
\While{~~$\mathbb{Q}_{b,j}^{(i)}\neq\emptyset$~~}{
$q\gets\mathbb{Q}_{b,j}^{(i)}(1)$:~Take the first (maximum) element of $\mathbb{Q}_{b,j}^{(i)}$ and assign it to $q$.\\
\SetAlgoVlined
\eIf{$q>s_{b,j}^{(i)}$}{
$\text{Insert}~\phi_{\mathrm{sort}}(1)~\text{into}~\mathbb{L}_{b,j}^{(i)}$.\\
$\mathbb{Q}_{b,j}^{(i)}\gets\mathbb{Q}_{b,j}^{(i)}-\mathbb{Q}_{b,j}^{(i)}(1)$.\\
Restart $k$ with $1$ and update the map $\phi_{\mathrm{sort}}:k\rightarrow c$.\\
Update $s_{b,j}^{(i)}$ using \eqref{eq:unspentAsset}.
}{
Break the while loop.
}}
If $\mathbb{L}_{b,j}^{(i)}=\emptyset$ then insert the reference classification label $l_{b,j}\in\mathbb{L}$ into $\mathbb{L}_{b,j}^{(i)}$.
\\
Return $\mathbb{L}_{b,j}^{(i)}$.}
\end{algorithm}

We denoted the vectorized reference labels, the fixed prior (win) probabilities, and the estimated posterior (belief) probabilities of the samples in the batch $\mathbb{T}_b\subseteq\mathbb{T}_{\mathrm{train}}$ with the $|\mathbb{T}_b|\times n_{\mathrm{clas}}$ matrices of $\mathbf{L}_b={[\mathbf{l}_{b,j}]}_j={[l_{b,j,c}]}_{j,c}$, $\mathbf{A}_b={[\mathbf{a}_{b,j}]}_j={[a_{b,j,c}]}_{j,c}$, and $\hat{\mathbf{P}}_b^{(i)}={[\hat{\mathbf{p}}_{b,j}^{(i)}]}_j={[\hat{p}_{b,j,c}^{(i)}]}_{j,c}$, respectively. Also, the allocation fractions estimated by the Kelly criterion for these samples formed a $|\mathbb{T}_b|\times n_{\mathrm{clas}}$ matrix denoted by $\hat{\mathbf{G}}_b^{(i)}={[\hat{\mathbf{g}}_{b,j}^{(i)}]}_j={[\hat{g}_{b,j,c}^{(i)}]}_{j,c}$.

In each iteration $i\in\{1,\cdots,n_{\mathrm{it}}\}$ of optimizing a discriminative neural network classifier, we first found the set of \textbf{candidate} classification labels $\mathbb{L}_{b,j}^{(i)}\subseteq\mathrm{L}$ for each sample (bettor) $v_{b,j}\in\mathbb{T}_b\subseteq\mathbb{T}_{\mathrm{train}}$. To this end, we proposed Algorithm \ref{Lopt} by using \eqref{eq:optFracClosed}, \eqref{eq:unspentAsset}, \eqref{eq:lnArgObj}, and \eqref{eq:ineqAsset}. Through this algorithm, the set of candidate labels $\mathbb{L}_{b,j}^{(i)}\subseteq\mathbb{L}$ got computed from the estimated posterior (belief) probabilities $\hat{\mathbf{p}}_{b,j}^{(i)}={\big[\hat{p}_{b,j,c}^{(i)}\in(0,1)\big]}_{c\in\mathbb{L}}$ and the fixed prior (win) probabilities $\mathbf{a}_{b,j}={\big[a_{b,j,c}\in(0,1)\big]}_{c\in\mathbb{L}}$ of the sample (bettor) $v_{b,j}\in\mathbb{T}_b\subseteq\mathbb{T}_{\mathrm{train}}$.

The set $\mathbb{L}_{b,j}^{(i)}\subseteq\mathbb{L}$ could contain multiple class labels or be empty. An empty set implied that the current posterior (belief) and the fixed prior (win) probabilities found no class label, even the reference label $l_{b,j}\in\mathbb{L}$, to be reliable enough for the optimization of the neural network classifier. This could result in no further update of the posterior (belief) probabilities in the following iterations. To avoid this standstill, at the end of the Algorithm \ref{Lopt}, if $\mathbb{L}_{b,j}^{(i)}=\emptyset$, then the reference label $l_{b,j}\in\mathbb{L}$ of the sample (bettor) $v_{b,j}\in\mathbb{T}_b\subseteq\mathbb{T}_{\mathrm{train}}$ got inserted into it.

By extending \eqref{eq:optAllocFrac} to all the samples in the batch $\mathbb{T}_b\subseteq\mathbb{T}_{\mathrm{train}}$, one obtained
\begin{equation}
\label{eq:objFuncKelly}
\mathbf{G}_b^{(i)}=\argmin_{\hat{\mathbf{G}}_b^{(i)}}~\underbrace{\frac{1}{|\mathbb{L}|\cdot|\mathbb{T}_b|}\sum_{j\in\mathbb{T}_b}\sum_{c\in\mathbb{L}}a_{b,j,c}\cdot\mathrm{ln}\bigg[1-\sum_{k\in\mathbb{L}}\hat{g}_{b,j,k}^{(i)}+\frac{\hat{g}_{b,j,c}^{(i)}}{\hat{p}_{b,j,c}^{(i)}}\bigg]}_{\mathcal{L}_{\mathrm{Kelly}}(\hat{\mathbf{G}}_b^{(i)})}.
\end{equation}

However, the optimum allocation fractions $\mathbf{G}_b^{(i)}={[\mathbf{g}_{b,j}^{(i)}]}_j={[g_{b,j,c}^{(i)}]}_{j,c}$ had a closed form solution given by \eqref{eq:optFracClosed}. This solution resulted in \eqref{eq:unspentAsset} and \eqref{eq:lnArgObj} and allowed to express
\begin{subequations}
\label{eq:minKellyObj}
\begin{align*}
&\min_{\hat{\mathbf{G}}_b^{(i)}}~\mathcal{L}_{\mathrm{Kelly}}(\hat{\mathbf{G}}_b^{(i)})=\mathcal{L}_{\mathrm{Kelly}}(\mathbf{G}_b^{(i)})=\frac{1}{|\mathbb{L}|\cdot|\mathbb{T}_b|}\sum_{j\in\mathbb{T}_b}\sum_{c\in\mathbb{L}}a_{b,j,c}\cdot\mathrm{ln}\bigg[s_{b,j}^{(i)}+\frac{g_{b,j,c}^{(i)}}{\hat{p}_{b,j,c}^{(i)}}\bigg]\tag{\ref{eq:minKellyObj}}\\
&=\frac{1}{|\mathbb{L}|\cdot|\mathbb{T}_b|}\sum_{j\in\mathbb{T}_b}\left[\sum_{c\in\mathbb{L}_{b,j}^{(i)}}a_{b,j,c}\cdot\mathrm{ln}\left[\frac{a_{b,j,c}}{\hat{p}_{b,j,c}^{(i)}}\right]+\left[\sum_{k\in\mathbb{L}-\mathbb{L}_{b,j}^{(i)}}a_{b,j,k}\right]\cdot\mathrm{ln}\left[\frac{\sum_{k\in\mathbb{L}-\mathbb{L}_{b,j}^{(i)}}a_{b,j,k}}{\sum_{k\in\mathbb{L}-\mathbb{L}_{b,j}^{(i)}}\hat{p}_{b,j,k}^{(i)}}\right]\right].
\end{align*}
\end{subequations}

\begin{table}[t!]
\begin{center}
\caption{Equivalence of the notations used in the objective functions of the active inference (left column) and the neural network optimization (right column).}
\resizebox{1.0\textwidth}{!}{%
\begin{tabular}{|c|l|}
\hline
$p(s|\pi)$&$l_{b,j,c}:~c^{\mathrm{th}}$ entry of the vectorized reference label of the sample $v_{b,j}\in\mathbb{T}_b\subseteq\mathbb{T}_{\mathrm{train}}$\\
$p(o)$&$a_{b,j,c}:$~prior (win) probability of the sample $v_{b,j}\in\mathbb{T}_b\subseteq\mathbb{T}_{\mathrm{train}}$\\
$q(o|\pi)$&$\hat{p}_{b,j,c}^{(i)}:$~estimated posterior (belief) probability of the sample $v_{b,j}\in\mathbb{T}_b\subseteq\mathbb{T}_{\mathrm{train}}$\\\hline\hline
\multicolumn{2}{|l|}{$\sum_{k\in\mathbb{L}-\mathbb{L}_{b,j}^{(i)}}a_{b,j,k}$:~collective prior of noncandidate classes of the sample $v_{b,j}\in\mathbb{T}_b\subseteq\mathbb{T}_{\mathrm{train}}$}\\
\multicolumn{2}{|l|}{$\sum_{k\in\mathbb{L}-\mathbb{L}_{b,j}^{(i)}}\hat{p}_{b,j,k}^{(i)}$:~collective posterior of noncandidate classes of the sample $v_{b,j}\in\mathbb{T}_b\subseteq\mathbb{T}_{\mathrm{train}}$}\\\hline
\end{tabular}}
\label{table:termsKellyEFE}
\end{center}
\end{table}

As given by \eqref{eq:crossEntrLoss}, the cross entropy loss for optimizing discriminative neural network classifiers was the variational free energy (VFE) of a retrospective active inference. That is,
\begin{equation}
\label{eq:crossEntrLoss2}
\mathcal{L}_{\mathrm{CE}}(\hat{\mathbf{P}}_b^{(i)},\mathbf{L}_b)=\frac{-1}{|\mathbb{L}|\cdot|\mathbb{T}_b|}\sum_{j\in\mathbb{T}_b}\sum_{c\in\mathbb{L}}l_{b,j,c}\cdot\mathrm{ln}\big(\hat{p}_{b,j,c}^{(i)}\big)~~\equiv~~-\sum_{s|\pi}p(s|\pi)\cdot\mathrm{ln}\big(q(o|\pi)\big).
\end{equation}

Also, the expected free energy (EFE) of a prospective active inference was given in \eqref{eq:objFuncEFE} as
\begin{equation}
\label{eq:objFuncEFE2}
\resizebox{0.9\textwidth}{!}{%
$\mathcal{L}_{\mathrm{EFE}}=\underbrace{\sum_{o}p(o)\cdot\Big[\mathrm{ln}\big(p(o)\big)-\mathrm{ln}\big(q(o|\pi)\big)\Big]}_{\mathrm{expected~complexity}}+\underbrace{\sum_{s|\pi}-p(s|\pi)\cdot\sum_{o|\pi}q(o|\pi)\cdot\mathrm{ln}\big(q(o|\pi)\big)}_{\mathrm{uncertainty}}.$%
}
\end{equation}

Our proposed Algorithm \ref{Lopt} for finding the candidate labels $\mathbb{L}_{b,j}^{(i)}$ aimed to minimize the objective function of the generalized Kelly criterion. This minimized function was given by \eqref{eq:minKellyObj}. A comparison of \eqref{eq:objFuncEFE2} and \eqref{eq:minKellyObj} with regard to \eqref{eq:crossEntrLoss2} revealed that the minimized objective of the Kelly criterion was the \textbf{expected~complexity} term of the EFE of a prospective active inference. That is, the objective function of the generalized Kelly criterion was a tight upper bound of the expected~complexity of the EFE. This equivalence got summarized in \autoref{table:termsKellyEFE} and implied that the \textbf{preferred observations} denoted by $o$ were realized through dividing $\mathbb{L}$ into candidate $\mathbb{L}_{b,j}^{(i)}$ and noncandidate classes $\mathbb{L}-\mathbb{L}_{b,j}^{(i)}$ and then handling the noncandidate classes altogether as one class. To this end, in \eqref{eq:minKellyObj}, the prior (win) probabilities of the noncandidate classes got summed together to form their collective prior (win) probability. Similarly, the estimated posterior (belief) probabilities of the noncandidate classes got summed together to form their collective posterior (belief) probability.

\begin{figure}[t!]
\begin{center}
\includegraphics[width=1.0\textwidth,height=19cm]{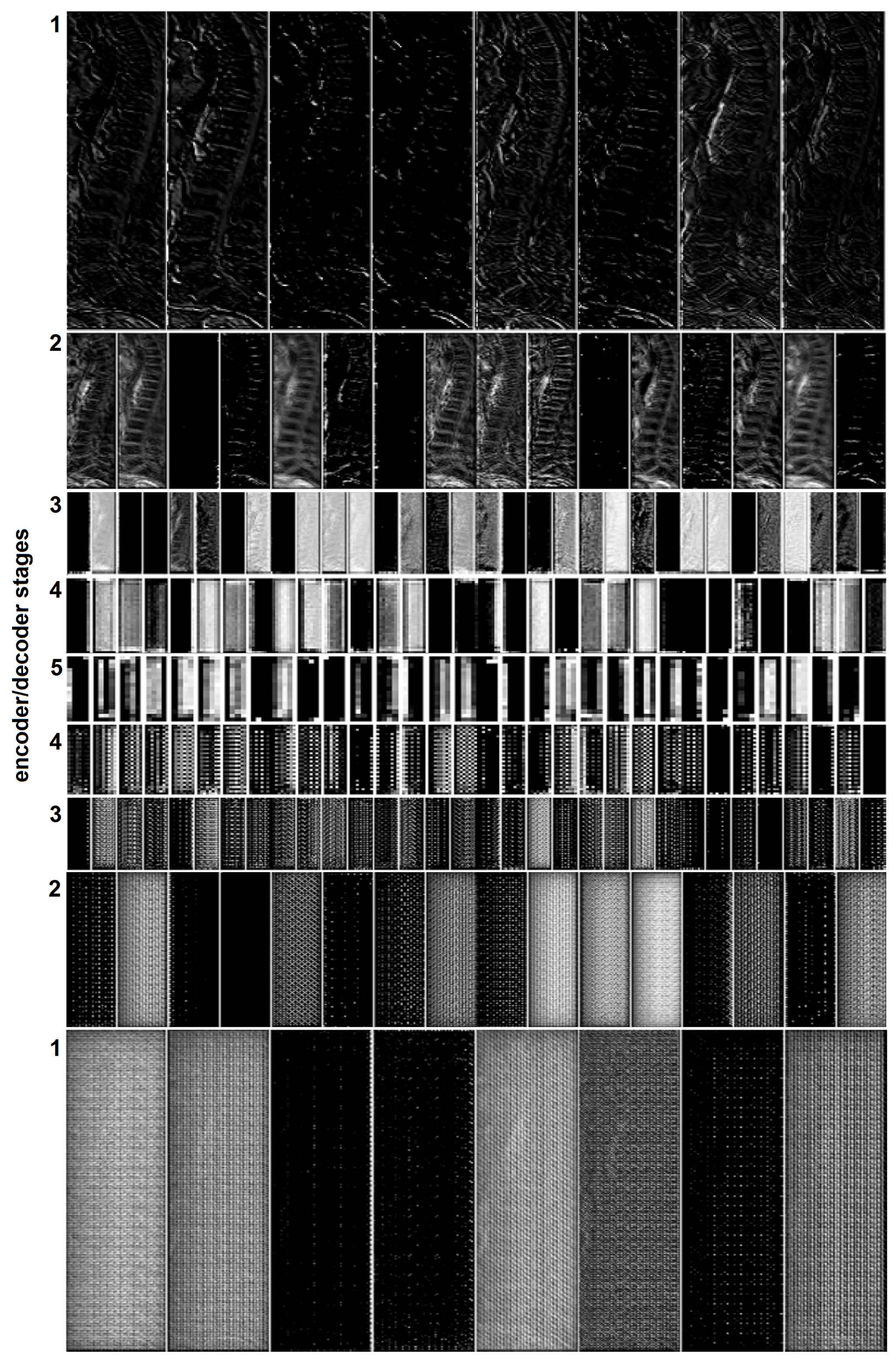}
\caption{Sagittal slices of the feature maps at the spatial regions enclosing the vertebral bodies and the intervertebral disks at the outputs of different encoder/decoder stages of the baseline architecture depicted in \autoref{fig:Vnet} after being optimized by the proposed objective function and its associated optimization process.}
\label{fig:FeatMapsVnet}
\end{center}
\end{figure}

The EFE in \eqref{eq:objFuncEFE2} was composed of an \textbf{expected~complexity} term plus an \textbf{uncertainty} term. As described in \autoref{ssec:activeInference}, the minimization of the expected~complexity was equivalent to the maximization of the reward. The reward maximization was also a goal of the Kelly criterion and could thus be partially fulfilled by finding the candidate labels through the proposed Algorithm \ref{Lopt}. To further maximize the reward, the expected~complexity should be minimized further. This was doable by having enough information or maximizing the information gain, i.e. minimizing the \textbf{uncertainty}. Accordingly, to optimize a discriminative neural network classifier, we proposed a novel objective function based on the EFE of a prospective active inference. The proposed function was given by
\begin{subequations}
\label{eq:propObjFunc}
\begin{align*}
&\mathcal{L}_{\mathrm{EFE}}(\hat{\mathbf{P}}_b^{(i)},\mathbf{A}_b,\mathbf{L}_b)=\underbrace{\frac{-1}{|\mathbb{L}|\cdot|\mathbb{T}_b|}\sum_{j\in\mathbb{T}_b}\sum_{c\in\mathbb{L}}{l_{b,j,c}\cdot\hat{p}_{b,j,c}^{(i)}\cdot\mathrm{ln}\big[\hat{p}_{b,j,c}^{(i)}\big]}}_{\mathrm{uncertainty}}~+\tag{\ref{eq:propObjFunc}}\\
&+\underbrace{\frac{1}{|\mathbb{L}|\cdot|\mathbb{T}_b|}\sum_{j\in\mathbb{T}_b}\Bigg[\sum_{c\in\mathbb{L}_{b,j}^{(i)}}a_{b,j,c}\cdot\mathrm{ln}\bigg[\frac{a_{b,j,c}}{\hat{p}_{b,j,c}^{(i)}}\bigg]+\bigg[\sum_{k\in\mathbb{L}-\mathbb{L}_{b,j}^{(i)}}a_{b,j,k}\bigg]\cdot\mathrm{ln}\bigg[\frac{\sum_{k\in\mathbb{L}-\mathbb{L}_{b,j}^{(i)}}a_{b,j,k}}{\sum_{k\in\mathbb{L}-\mathbb{L}_{b,j}^{(i)}}\hat{p}_{b,j,k}^{(i)}}\bigg]\Bigg]}_{\mathrm{expected~complexity}}.
\end{align*}
\end{subequations}
This function was reversible and differentiable with respect to the posteriors $\hat{\mathbf{P}}_b^{(i)}$. As given by \eqref{eq:softmaxPostNet}, these posteriors were generated by applying the Softmax function to the network's outputs $\mathbf{Z}_b^{(i)}={[\mathbf{z}_{b,j}^{(i)}]}_j={[z_{b,j,c}^{(i)}]}_{j,c}$. Thus, the proposed function was also differentiable with respect to the $\mathbf{Z}_b^{(i)}$ and the outputs of every layer. As described in \autoref{ssec:optNeuralNetClass}, these allowed to minimize it by a gradient descent optimizer with backpropagation.

We preceded the minimization of \eqref{eq:propObjFunc} with a partial minimization of its expected~complexity term by finding the candidate classification labels $\mathbb{L}_{b,j}^{(i)}$ of each sample (bettor) $v_{b,j}\in\mathbb{T}_b\subseteq\mathbb{T}_{\mathrm{train}}$ through the Algorithm \ref{Lopt} proposed based on the Kelly criterion.

Accordingly, in each iteration $i\in\{1,\cdots,n_{\mathrm{it}}\}$ of our proposed optimization process, every sample $v_j\in\mathbb{T}_b\subseteq\mathbb{T}_{\mathrm{train}}$ got passed through the network to estimate its classification posteriors $\hat{\mathbf{P}}_b^{(i)}={\big[\hat{\mathbf{p}}_{b,j}^{(i)}\in(0,1)\big]}_j={[\hat{p}_{b,j,c}^{(i)}]}_{j,c}$. From these posteriors and the fixed priors $\mathbf{a}_{b,j}={\big[a_{b,j,c}\in(0,1)\big]}_{c\in\mathbb{L}}$ of the sample, its candidate classification labels $\mathbb{L}_{b,j}^{(i)}\subseteq\mathbb{L}$ got computed by using the proposed Algorithm \ref{Lopt}. Then, the loss at the last network's layer got obtained by inputting the posteriors, the priors, and the candidate labels of the samples into the proposed function in \eqref{eq:propObjFunc}. By propagating this loss from the last layer to the first layer, the loss of every layer got obtained. Then, the gradient (first derivative) of each layer's loss got calculated with respect to its outputs. The product of these layerwise gradients got used by the gradient descent optimizer to update the network's parameters.

In an image segmentation task, each sample $v_{b,j}\in\mathbb{T}_b\subseteq\mathbb{T}_{\mathrm{train}}$ was an image patch processed by a network's layer. In our baseline architecture described in \autoref{ssec:baselineArchi}, each network's layer processed samples (patches) of a certain spatial resolution. The multiresolution hierarchy of the network was the result of downsampling and upsampling each volumetric fat-water image through convolutional and deconvolutional layers, respectively. For sake of simplicity, we omitted the resolution specifying indices from the samples' notations.

\autoref{fig:FeatMapsVnet} shows sagittal slices of the feature maps at the spatial regions enclosing the vertebral bodies and the intervertebral disks at the outputs of different encoder/decoder stages of the baseline architecture depicted in \autoref{fig:Vnet} after being optimized by the proposed objective function and its associated optimization process.

\section{Network's Parameters and Their Optimization}
\label{sec:optBaseArchi}
Our proposed algorithm for finding the candidate labels and our proposed objective function for optimizing a discriminative neural network classifier got integrated into a mini-batch-based gradient descent optimizer with backpropagation by using the process proposed in \autoref{sec:PropOptProc}. This process got evaluated against a similar process incorporating a representative of the cross entropy-based losses or a representative of the metric-based losses introduced in \autoref{ssec:CommObjsBotlncks}. The representative of the cross entropy-based losses was the weighted focal loss. This loss comprised of a modulating factor and a weighting mechanism to alleviate classification biases towards the dominant classes of the training samples. The representative of the metric-based losses was the Lov\'{a}sz-Softmax loss. Besides being smooth and differentiable, to the best of our knowledge, this loss was the only convex loss among the metric-based losses.

Accordingly, the evaluated losses were
\begin{enumerate}[label={(\arabic*)},leftmargin=*]
\item the proposed objective function (Po) given by \eqref{eq:propObjFunc}
\item the weighted focal loss (Fo) given by \eqref{eq:wghtFocLoss}
\item the Lov\'{a}sz-Softmax loss (Lo) given by \eqref{eq:lovaszSoftMult}.
\end{enumerate}
These evaluations were on an end-to-end optimization of the baseline architecture described in \autoref{ssec:baselineArchi}. For each case, the baseline architecture was once used without attention gates (Na) as depicted in \autoref{fig:Vnet} and once used with the attention gates (At) as depicted in \autoref{fig:VnetAttention}. Also, for (2) and (3) each training sample was accompanied by its reference (ground truth) label to fulfill the supervised nature of these objective functions. However, our proposed algorithm for finding the candidate labels and our proposed objective function got evaluated according to a fully supervised, a semi-supervised, and an unsupervised approach. These resulted in the training samples being
\begin{enumerate}[label={(\arabic*)},leftmargin=*]
\item accompanied by their reference labels and their priors (GrPr)~$\rightarrow$~fully supervised
\item only accompanied by their reference labels (GrNp)~$\rightarrow$~semi-supervised
\item only accompanied by their priors (NgPr)~$\rightarrow$~semi-supervised
\item accompanied by neither their reference labels nor their priors (NgNp)~$\rightarrow$~unsupervised.
\end{enumerate}
For the cases with the priors, the prior probabilities of the training samples could be computed by a multiatlas registration. If no prior probabilities were provided at the time of optimization (training), then uniform priors got assumed. If the reference (ground truth) labels of the training samples $\mathbb{T}_{\mathrm{train}}$ were provided at the time of optimization (training), then for each sample $v_{b,j}\in\mathbb{T}_b\subseteq\mathbb{T}_{\mathrm{train}}$ the vectorized reference label $\mathbf{l}_{b,j}$ was the one-hot-encoding of its reference label $l_{b,j}\in\mathbb{L}$ and was given by \eqref{eq:refLabelOneHot}. If the reference labels of the training samples $\mathbb{T}_{\mathrm{train}}$ were not provided at the time of optimization, then for each sample $v_{b,j}\in\mathbb{T}_b\subseteq\mathbb{T}_{\mathrm{train}}$ the vector $\mathbf{l}_{b,j}$ was uniform and given by \eqref{eq:refLabelUniform}.

For each evaluation case, the main parameters and the hyperparameters of the baseline architecture got trained (optimized) to automatically segment $n_{\mathrm{clas}}=|\mathbb{L}|=8$ classes of vertebral bodies (VBs), intervertebral disks (IVDs), psoas major (PM) and quadratus lumborum (QL) muscles, epicardial adipose tissues (EpAT), pericardial adipose tissues (PeAT), cardiac perivascular adipose tissues (PvAT), and background on each volumetric fat-water image. To this end, the volumetric fat-water images got divided into a training and a test set. The training set formed the samples set $\mathbb{T}_{\mathrm{train}}$ and got used to optimize the main parameters and the hyperparameters of the baseline architecture by each method. The test set formed the samples set $\mathbb{T}_{\mathrm{test}}$ and got used to evaluate the classification performance of the baseline architecture after being fully optimized by each method. The training set was composed of samples accompanied by their reference labels and priors. The test set was composed of samples accompanied by their reference labels. The reference labels of the test samples were not fed to the neural network. They were rather compared against the corresponding labels predicted by the network to evaluate the classification performance of the network. The predicted label of each sample was the index of its maximum classification posterior estimated by the network.

The main parameters of the baseline architecture included the weights and the biases of the convolutional and deconvolutional layers, the leakage coefficient $a_{\mathrm{prelu}}\in\mathbb{R}_{\geq 0}$ of every nonlinear PReLU activation, and the means and variances of the (instance) normalizers introduced in page~\pageref{instanceNorm}. Prior to the optimization of the main parameters, they should be initialized. This initialization was extremely important for the weights of the convolutional and deconvolutional layers of a residual network of several layers and thus different paths of signal propagation. Without a proper weight initialization, some parts of the network might have excessive activations and thus produce stronger gradients while some other parts might produce weaker gradients and thus get optimized less. To avoid this, a random initialization of the weights with the aim of breaking symmetries and making each feature map of a unit variance was suggested. For this, the weights were drawn from a certain distribution. In networks with nonlinear Sigmoid or hyperbolic tangent activations as well as linear activations, the proper initializations of the weights of every layer were random numbers drawn from a uniform distribution in the range $[-\sqrt{6/(n_{\mathrm{in}}+n_{\mathrm{out}})},~\sqrt{6/(n_{\mathrm{in}}+n_{\mathrm{out}})}]$ with $n_{\mathrm{in}}$ being the number of incoming network connections (fan-in) and $n_{\mathrm{out}}$ being the number of outgoing network connections (fan-out) of the layer. This type of initialization was called a \textit{Glorot} or a \textit{Xavier} initialization and was shown to be improper for networks involving nonlinear rectified linear units, including the PReLU, as their activations \cite{Glorot2010}.
For these networks, like our baseline architecture, the proper initializations of the weights of every convolutional/deconvolutional layer were random numbers drawn from a Gaussian distribution with a mean of 0 and a standard deviation of $\sqrt{2/n_{\mathrm{in}}}$ \cite{He2015,Ronneberger2015}. For a convolutional layer of a kernel size of $5\times5\times5$, $16$ input feature maps, and $32$ output feature maps, the number of incoming network connections (fan-in) was $5\times5\times5\times16=2000$ and the number of outgoing network connections (fan-out) was $32$. The biases of every convolutional/deconvolutional layer were initialized to 0. The leakage coefficient of every nonlinear PReLU activation got initialized to $0.15$ to allow a small leakage of negative inputs. The means and the variances of the (instance) normalizers got initialized to 0 and 1 respectively.

The hyperparameters of the baseline architecture and their discretized values were
\begin{itemize}[leftmargin=*]
\setlength\itemsep{-0.2em}
\item number of convolutional/deconvolutional layers $n_s\in\{1,2,\cdots,5\}$ of the $s^{\mathrm{th}}$ encoder/decoder stage of the V-net of the baseline architecture
\item Dropout's retention probability $p_s\in\{0.1,0.2,\cdots,0.9\}$ of the perceptrons (nodes) of the $s^{\mathrm{th}}$ encoder/decoder stage of the V-net of the baseline architecture.
\end{itemize}

\begin{figure}[t!]
\begin{center}
\includegraphics[width=1.0\textwidth,height=21cm]{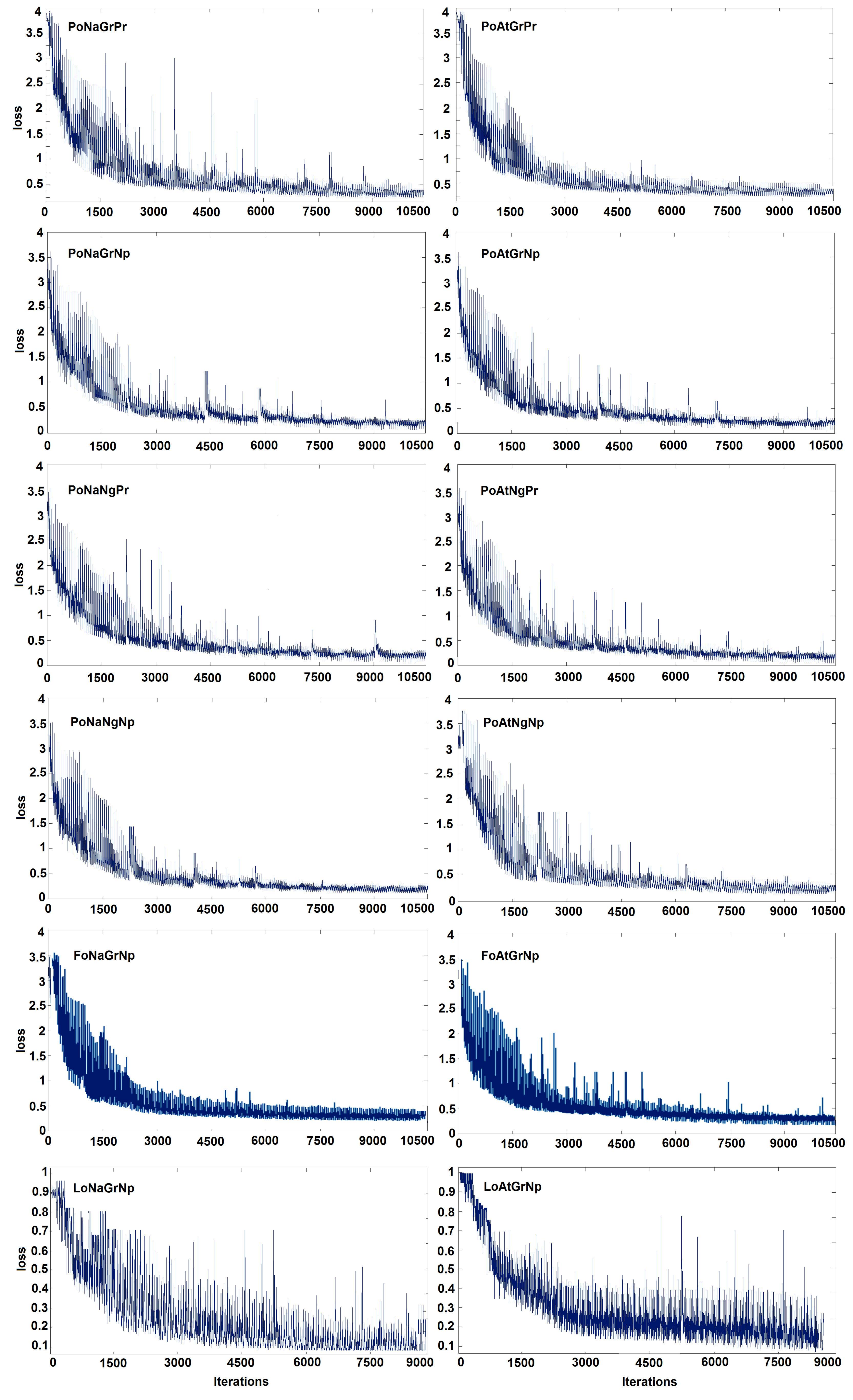}
\caption{Convergence patterns of different evaluation cases with each case optimizing its main parameters with the best performing hyperparameters.}
\label{fig:LossConvergVnet}
\end{center}
\end{figure}

To optimize the main parameters and the hyperparameters of the baseline architecture by each method, a random search over the discretized hyperparameter values and a 5-fold cross validation were conducted. To this end, the training set got divided into 5 subsets. Then, for each method, in each optimization trial, a set of hyperparameter values got randomly selected. With these hyperparameter values, 5 times training and validation got performed according to the 5-fold cross validation. In each fold, the main parameters of the baseline architecture got optimized on 4 subsets by using a mini-batch-based gradient descent optimizer with backpropagation. The gradient descent optimizer was the Adam optimizer described in \autoref{ssec:optNeuralNetClass}. The resulting network model got then evaluated on the remaining (validation) subset by calculating the \textit{precision} and the \textit{recall} metrics for each of the $n_{\mathrm{clas}}-1=8-1=7$ foreground classes against the rest of the classes. This way, for the selected hyperparameter values, at the end of the 5-fold cross validation, 5 network models and 7 \textit{precision} and 7 \textit{recall} values per network model were obtained. For each model, the 7 \textit{precision} and the 7 \textit{recall} values got averaged. Then, for the selected hyperparameter values, the model of maximum averaged \textit{precision} and \textit{recall} was the best performing model. The optimization trials continued by randomly selecting another set of hyperparameter values until the best performing model resulted from the current hyperparameter values could not exceed the averaged \textit{precision} and \textit{recall} values of any of the best models in the last 50 trials.
The \textit{precision} and \textit{recall} metrics were selected due to their robustness against the imbalanced class-sample distributions. Moreover, the aforementioned cross validation aimed to reduce the impacts of the randomized initialization of the main parameters on the resulting network models. The 5 folds were selected with regard to the maximum size of the baseline architecture and the sufficiency of the number of training and validation samples for the optimization and evaluation in each fold, respectively.
The above process was done by using the tools provided in the distributed asynchronous hyperparameter optimization (Hyperopt) library in Python \cite{Bergstra2015}. For the hyperparameter selection, in addition to the randomization, this library provided a tree of Parzen estimators (TPE) and its adaptive variant. The TPE was more appropriate for belief neural networks of undirected graph topology than the feed-forward networks like our baseline architecture \cite{Bergstra2011,Bergstra2012}.

\begin{table}[t!]
\begin{center}
\caption{Optimized hyperparameters and the overall time [hours] of optimizing the main parameters and the hyperparameters for each evaluation case.}
\vspace{2mm}
\resizebox{0.7\textwidth}{!}{%
\begin{tabular}{|ll|cccccc|}
\hline
&&\multicolumn{6}{c|}{\textbf{Evaluation Case}}\\
&&PoNaGrPr&PoNaGrNp&PoNaNgPr&PoNaNgNp&PoAtGrPr&PoAtGrNp\\\hline
\parbox[t]{0mm}{\multirow{11}{*}{\rotatebox[origin=c]{90}{\textbf{Hyperparameters}}}}
&$n_1$&2&2&2&3&1&1\\
&$n_2$&2&3&2&4&2&2\\
&$n_3$&3&3&4&4&2&2\\
&$n_4$&3&4&4&5&3&3\\
&$n_5$&3&4&5&5&3&4\\
&$p_1$&0.9&0.8&0.8&0.7&0.9&0.9\\
&$p_2$&0.8&0.7&0.7&0.7&0.7&0.8\\
&$p_3$&0.7&0.7&0.7&0.7&0.7&0.7\\
&$p_4$&0.7&0.7&0.7&0.6&0.6&0.7\\
&$p_5$&0.6&0.6&0.6&0.6&0.5&0.6\\
&time&83&90&95&107&82&85\\
\hline\hline
&&\multicolumn{6}{c|}{\textbf{Evaluation Case}}\\
&&PoAtNgPr&PoAtNgNp&FoNaGrNp&FoAtGrNp&LoNaGrNp&LoAtGrNp\\\hline
\parbox[t]{0mm}{\multirow{11}{*}{\rotatebox[origin=c]{90}{\textbf{Hyperparameters}}}}
&$n_1$&1&2&2&2&1&1\\
&$n_2$&2&3&3&2&2&2\\
&$n_3$&3&4&4&3&3&2\\
&$n_4$&3&4&4&3&4&3\\
&$n_5$&4&5&4&4&4&3\\
&$p_1$&0.9&0.9&0.9&0.9&0.9&0.9\\
&$p_2$&0.8&0.8&0.8&0.8&0.8&0.8\\
&$p_3$&0.7&0.8&0.8&0.7&0.7&0.7\\
&$p_4$&0.6&0.7&0.7&0.6&0.7&0.6\\
&$p_5$&0.5&0.6&0.6&0.5&0.5&0.5\\
&time&88&97&84&82&79&77\\
\hline
\end{tabular}}
\label{table:hypOptNeurNet}
\end{center}
\end{table}

The evaluated objective functions and the Adam-based gradient descent optimizer involved following fixed parameters:
\begin{itemize}[leftmargin=*]
\item N=$|\mathbb{T}_b|=2$: As explained in page~\pageref{numBatches}, due to the memory limitations of the used GPU, only 2 volumetric fat-water images were included in each mini-batch.
\item $\gamma_{\mathrm{mod}}=2$: Modulating factor of the focal loss given by \eqref{eq:focalLoss}.
\item $\alpha_{\mathrm{lr}}=0.001$: Learning rate (step size) of the gradient descent optimizer defined in \eqref{eq:updateMainParams}. This learning rate did not need to be adapted manually as the Adam optimizer automatically changed the effective learning rate by the ratio of the exponential moving average of the first moment to the exponential moving average of the second moment.
\item $\beta_{\mathrm{fm}}=0.90$: Decay rate of the estimated first moments.
\item $\beta_{\mathrm{sm}}=0.99$: Decay rate of the estimated second moments.
\item $\boldsymbol{m}^{(0)}=\mathbf{0}$: Initial first moments.
\item $\boldsymbol{v}^{(0)}=\mathbf{0}$: Initial second moments.
\end{itemize}
The number of iterations $n_{\mathrm{it}}\in\{10,\cdots,15000\}$ was determined according to an early stopping criterion. That is, when the exponential moving average of the validation error (loss) was not improved within the last 100 iterations, then the optimization got stopped.

\autoref{fig:LossConvergVnet} shows convergence patterns of different evaluation cases with each case optimizing its main parameters with the best performing hyperparameters.

The aforementioned optimizations were conducted on 4 NVIDIA TITAN X\textsuperscript{\textregistered} GPUs of 12 GB memory each and by using a memory efficient cuDNN3 implementation of the convolutional/deconvolutional layers and the TensorFlow\textsuperscript{TM} library of version 2.3 \cite{Abadi2016}.

\autoref{table:hypOptNeurNet} shows the optimized hyperparameters and the overall time of optimizing the main parameters and the hyperparameters for each evaluation case. After the optimizations, an automatic segmentation of the $n_{\mathrm{clas}}=8$ classes on an unseen volumetric fat-water image took around 3 seconds for each evaluation case on the GPUs used for the optimizations.
\begin{singlespace}
{\footnotesize
\bibliography{ArtklBookMisc,IEEEabrvIndexMedicus,ProcLong}}

\begin{thebibliography}{Badrinarayanan 2016x}

\bibitem[{Abadi} 2016]{Abadi2016}
Mart{\'\i}n {Abadi}, Ashish {Agarwal}, Paul {Barham}, Eugene {Brevdo}, Zhifeng
  {Chen}, Craig {Citro}, Greg~S. {Corrado}, Andy {Davis}, Jeffrey {Dean},
  Matthieu {Devin}, Sanjay {Ghemawat}, Ian {Goodfellow}, Andrew {Harp},
  Geoffrey {Irving}, Michael {Isard}, Yangqing {Jia}, Rafal {Jozefowicz},
  Lukasz {Kaiser}, Manjunath {Kudlur}, Josh {Levenberg}, Dan {Mane}, Rajat
  {Monga}, Sherry {Moore}, Derek {Murray}, Chris {Olah}, Mike {Schuster},
  Jonathon {Shlens}, Benoit {Steiner}, Ilya {Sutskever}, Kunal {Talwar}, Paul
  {Tucker}, Vincent {Vanhoucke}, Vijay {Vasudevan}, Fernanda {Viegas}, Oriol
  {Vinyals}, Pete {Warden}, Martin {Wattenberg}, Martin {Wicke}, Yuan {Yu} and
  Xiaoqiang {Zheng}, \emph{{TensorFlow: Large-scale machine learning on
  heterogeneous distributed systems}}. \emph{arXiv e-prints}, March 2016,
  https://www.tensorflow.org/.

\bibitem[Alom 2019]{Alom2019}
Md~Zahangir Alom, Chris Yakopcic, Mahmudul Hasan, Tarek~M. Taha and Vijayan~K.
  Asari, \emph{{Recurrent residual U-Net for medical image segmentation}}.
  \emph{J Med Imaging}, Volume~6, 2019.

\bibitem[Bach 2013]{Bach2013}
Francis Bach, \emph{{Learning with submodular functions: A convex optimization
  perspective}}. \emph{Found Trends Mach Learn}, Volume~6, Pages 145--373,
  2013.

\bibitem[Badrinarayanan 2016]{Badrinarayanan2016}
V.~Badrinarayanan, A.~Kendall and R.~Cipolla, \emph{{SegNet: A deep
  convolutional encoder-decoder architecture for image segmentation}}.
  \emph{ArXiv e-prints}, 2016.

\bibitem[Bergstra 2011]{Bergstra2011}
James Bergstra, R\'{e}mi Bardenet, Yoshua Bengio and Bal\'{a}zs K\'{e}gl,
  \emph{{Algorithms for hyperparameter optimization}}. In \emph{Proceedings of
  Advances in Neural Information Processing Systems}, Pages 2546--2554, 2011.

\bibitem[Bergstra 2012]{Bergstra2012}
James Bergstra and Yoshua Bengio, \emph{{Random search for hyperparameter
  optimization}}. \emph{J Mach Learn Res}, Volume~13, Pages 281--305, feb 2012.

\bibitem[Bergstra 2015]{Bergstra2015}
James Bergstra, Brent Komer, Chris Eliasmith, Dan Yamins and David~D Cox,
  \emph{{Hyperopt: a Python library for model selection and hyperparameter
  optimization}}. \emph{Comput Sci Discov}, Volume~8, Page 014008, 2015,
  https://github.com/hyperopt/hyperopt.

\bibitem[Berman 2018]{Berman2018}
Maxim Berman, Amal~Rannen Triki and Matthew~B. Blaschko, \emph{{The
  Lovasz-Softmax loss: A tractable surrogate for the optimization of the
  intersection-over-union measure in neural networks}}. In \emph{Proceedings of
  the IEEE Conference on Computer Vision and Pattern Recognition}, Pages
  4413--4421, 2018.

\bibitem[Bertels 2019]{Bertels2019}
Jeroen Bertels, Tom Eelbode, Maxim Berman, Dirk Vandermeulen, Frederik Maes,
  Raf Bisschops and Matthew~B. Blaschko, \emph{{Optimizing the Dice score and
  Jaccard index for medical image segmentation: Theory and practice}}. In
  \emph{Proceedings of the International Conference on Medical Image Computing
  and Computer-Assisted Intervention}, Pages 92--100, 2019.

\bibitem[Boyd 2004]{Boyd2004}
S.P. Boyd and L.~Vandenberghe, \emph{Convex Optimization}. Cambridge University
  Press, 2004.

\bibitem[Carr 2022]{Carr2022}
Steven Carr, Nils Jansen and Ufuk Topcu, \emph{{Task-aware verifiable RNN-based
  policies for partially observable Markov decision processes}}. \emph{J Artif
  Intell Res}, Volume~72, Pages 819--847, jan 2022.

\bibitem[Chen 2022]{Chen2022}
Tzu-Hsuan Chen and Tian~Sheuan Chang, \emph{{RangeSeg: Range-Aware real time
  segmentation of 3D LiDAR point clouds}}. \emph{IEEE Trans Intell Veh},
  Volume~7, Pages 93--101, 2022.

\bibitem[{\c{C}}i{\c{c}}ek 2016]{Cicek2016}
{\"O}zg{\"u}n {\c{C}}i{\c{c}}ek, Ahmed Abdulkadir, Soeren~S. Lienkamp, Thomas
  Brox and Olaf Ronneberger, \emph{{3D U-Net: Learning dense volumetric
  segmentation from sparse annotation}}. In \emph{Proceedings of the
  International Conference on Medical Image Computing and Computer-Assisted
  Intervention}, Pages 424--432, 2016.

\bibitem[Cui 2019]{Cui2019}
Yin Cui, Menglin Jia, Tsung-Yi Lin, Yang Song and Serge Belongie,
  \emph{{Class-balanced loss based on effective number of samples}}. In
  \emph{Proceedings of the IEEE Conference on Computer Vision and Pattern
  Recognition}, Pages 9260--9269, 2019.

\bibitem[Dean 2012]{Dean2012}
Jeffrey Dean, Greg Corrado, Rajat Monga, Kai Chen, Matthieu Devin, Mark Mao,
  Marc Ranzato, Andrew Senior, Paul Tucker, Ke~Yang, Quoc~V. Le and Andrew~Y.
  Ng, \emph{{Large scale distributed deep networks}}. In \emph{Proceedings of
  Advances in Neural Information Processing Systems}, Pages 1223--1231, 2012.

\bibitem[Feldman 2010]{Feldman2010}
Harriet Feldman and Karl Friston, \emph{{Attention, uncertainty, and
  free-energy}}. \emph{Front Hum Neurosci}, Volume~4, 2010.

\bibitem[Friston 2019]{Friston2019}
Karl Friston, \emph{{A free energy principle for a particular physics}}.
  \emph{ArXiv e-prints}, 2019.

\bibitem[Fujishige 1991]{Fujishige1991}
S.~Fujishige, \emph{{Submodular functions and optimization}}. ISSN, Elsevier
  Science, 1991.

\bibitem[Gal 2015]{Gal2015}
Yarin Gal and Zoubin Ghahramani, \emph{{Dropout as a Bayesian approximation:
  Representing model uncertainty in deep learning}}. \emph{ArXiv e-prints},
  2015.

\bibitem[Glorot 2010]{Glorot2010}
Xavier Glorot and Yoshua Bengio, \emph{{Understanding the difficulty of
  training deep feed-forward neural networks}}. In \emph{Proceedings of the
  International Conference on Artificial Intelligence and Statistics},
  Volume~9, Pages 249--256, 2010.

\bibitem[Goodfellow 2016]{Goodfellow2016}
Ian Goodfellow, Yoshua Bengio and Aaron Courville, \emph{Deep Learning}. MIT
  Press, 2016.

\bibitem[He 2015]{He2015}
Kaiming He, Xiangyu Zhang, Shaoqing Ren and Jian Sun, \emph{{Delving deep into
  rectifiers: Surpassing human-level performance on ImageNet classification}}.
  In \emph{Proceedings of the IEEE International Conference on Computer
  Vision}, Pages 1026--1034, 2015.

\bibitem[He 2016{a}]{He2016}
Kaiming He, Xiangyu Zhang, Shaoqing Ren and Jian Sun, \emph{{Deep residual
  learning for image recognition}}. In \emph{Proceedings of the IEEE Conference
  on Computer Vision and Pattern Recognition}, Pages 770--778, 2016.

\bibitem[He 2016{b}]{Kaiming2016}
Kaiming He, Xiangyu Zhang, Shaoqing Ren and Jian Sun, \emph{{Identity mappings
  in deep residual networks}}. In \emph{Proceedings of the European Conference
  on Computer Vision}, Pages 630--645, 2016.

\bibitem[Ioffe 2015]{Ioffe2015}
Sergey Ioffe and Christian Szegedy, \emph{{Batch normalization: Accelerating
  deep network training by reducing internal covariate shift}}. In
  \emph{Proceedings of the International Conference on Machine Learning},
  Volume~37, Pages 448--456, 2015.

\bibitem[Isensee 2018]{Isensee2018}
Fabian Isensee, Philipp Kickingereder, Wolfgang Wick, Martin Bendszus and
  Klaus~H. Maier-Hein, \emph{{No New-Net}}. In \emph{Proceedings of the
  International Workshop on Bainlesion: Glioma, Multiple Sclerosis, Stroke and
  Traumatic Brain Injuries}, Pages 234--244, 2018.

\bibitem[Jadon 2020]{Jadon2020}
Shruti Jadon, \emph{{A survey of loss functions for semantic segmentation}}. In
  \emph{Proceedings of the IEEE Conference on Computational Intelligence in
  Bioinformatics and Computational Biology}, Pages 1--7, 2020.

\bibitem[Jospin 2022]{Jospin2022}
Laurent~Valentin Jospin, Wray~L. Buntine, Farid Boussa{\"{\i}}d, Hamid Laga and
  Mohammed Bennamoun, \emph{{Hands-on Bayesian neural networks - A tutorial for
  deep learning users}}. \emph{ArXiv e-prints}, 2022.

\bibitem[Kelly 1956]{Kelly1956}
J.~L. Kelly, \emph{{A new interpretation of information rate}}. \emph{Bell Syst
  Tech J}, Volume~35, Pages 917--926, 1956.

\bibitem[Kingma 2015]{Kingma2015}
Diederik~P. Kingma and Jimmy Ba, \emph{{Adam: A method for stochastic
  optimization}}. In \emph{Proceedings of International Conference on Learning
  Representations}, Edited by Yoshua Bengio and Yann LeCun, 2015.

\bibitem[Kullback 1951]{Kullback1951}
S.~Kullback and R.~A. Leibler, \emph{{On information and sufficiency}}.
  \emph{Ann Math Statist}, Pages 79--86, 1951.

\bibitem[Li 2022]{Sijia2022}
Sijia Li, Furkat Sultonov, Qingshan Ye, Yong Bai, Jun-Hyun Park, Chilsig Yang,
  Minseok Song, Sungwoo Koo and Jae-Mo Kang, \emph{{TA-Unet: Integrating
  triplet attention module for drivable road region segmentation}}.
  \emph{Sensors}, Volume~22, 2022.

\bibitem[Lin 2018]{Lin2018}
T.~Lin, P.~Goyal, R.~Girshick, K.~He and P.~Dollar, \emph{{Focal loss for dense
  object detection}}. \emph{IEEE Trans Pattern Anal Mach Intell}, 2018.

\bibitem[Lo 2021]{Lo2021}
Justin Lo, Jillian Cardinell, Alejo Costanzo and Dafna Sussman, \emph{{Medical
  augmentation (Med-Aug) for optimal data augmentation in medical deep learning
  networks}}. \emph{Sensors}, Volume~21, 2021.

\bibitem[Lov{\'a}sz 1983]{Lovasz1983}
L.~Lov{\'a}sz, \emph{Submodular functions and convexity}. In \emph{Mathematical
  Programming The State of the Art}, Pages 235--257, Springer Berlin
  Heidelberg, 1983.

\bibitem[Maier 2019]{Maier2019}
Andreas~K. Maier, Christopher Syben, Bernhard Stimpel, Tobias W{\"u}rfl, Mathis
  Hoffmann, Frank Schebesch, Weilin Fu, Leonid Mill, Lasse Kling and Silke~H.
  Christiansen, \emph{{Learning with known operators reduces maximum training
  error bounds}}. \emph{Nat Mach Intell}, Volume~1, Pages 373--380, 2019.

\bibitem[McMillan 1956]{McMillan1956}
B.~McMillan, \emph{{Two inequalities implied by unique decipherability}}.
  \emph{IEEE Trans Inf Theory}, Pages 115--116, 1956.

\bibitem[Milletari 2016]{Milletari2016}
F.~Milletari, N.~Navab and S.~Ahmadi, \emph{{V-Net: Fully Convolutional Neural
  Networks for Volumetric Medical Image Segmentation}}. In \emph{International
  Conference on 3D Vision}, Pages 565--571, 2016.

\bibitem[Mubashar 2022]{Mehreen2022}
Mehreen Mubashar, Hazrat Ali, Christer Gr\"{o}nlund and Shoaib Azmat,
  \emph{{R2U++: A multiscale recurrent residual U-Net with dense skip
  connections for medical image segmentation}}. \emph{Neural Comput Appl},
  Volume~34, Pages 17723--17739, 2022.

\bibitem[Oktay 2018]{Oktay2018}
Ozan Oktay, Jo~Schlemper, Loic Le~Folgoc, Matthew Lee, Mattias Heinrich,
  Kazunari Misawa, Kensaku Mori, Steven McDonagh, Nils Y~Hammerla, Bernhard
  Kainz, Ben Glocker and Daniel Rueckert, \emph{{Attention U-Net: Learning
  where to look for the pancreas}}. \emph{ArXiv e-prints}, 2018.

\bibitem[Rakhlin 2018]{Rakhlin2018}
Alexander Rakhlin, Alex Davydow and Sergey Nikolenko, \emph{{Land cover
  classification from satellite imagery with U-Net and Lov\'{a}sz-Softmax
  loss}}. In \emph{Proceedings of the IEEE Conference on Computer Vision and
  Pattern Recognition Workshops}, Pages 257--2574, 2018.

\bibitem[Ronneberger 2015]{Ronneberger2015}
Olaf Ronneberger, Philipp Fischer and Thomas Brox, \emph{{U-Net: Convolutional
  networks for biomedical image segmentation}}. In \emph{Proceedings of the
  International Conference on Medical Image Computing and Computer-Assisted
  Intervention}, Pages 234--241, 2015.

\bibitem[Ruder 2016]{Ruder2016}
Sebastian Ruder, \emph{{An overview of gradient descent optimization
  algorithms}}. \emph{ArXiv e-prints}, 2016.

\bibitem[Smith 2022]{Smith2022}
Ryan Smith, Karl~J Friston and Christopher~J Whyte, \emph{{A step-by-step
  tutorial on active inference and its application to empirical data}}. \emph{J
  Math Psychol}, Volume 107, 2022.

\bibitem[Smoczynski 2010]{Smoczynski2010}
Peter Smoczynski and Dave Tomkins, \emph{An explicit solution to the problem of
  optimizing the allocations of a Bettor's wealth when wagering on horse
  races}. \emph{Math Sci}, Volume~35, Pages 10--17, 2010.

\bibitem[Srivastava 2014]{Srivastava2014}
Nitish Srivastava, Geoffrey Hinton, Alex Krizhevsky, Ilya Sutskever and Ruslan
  Salakhutdinov, \emph{{Dropout: A simple way to prevent neural networks from
  overfitting}}. \emph{J Mach Learn Res}, Volume~15, Pages 1929--1958, 2014.

\bibitem[Tatbul 2018]{Tatbul2018}
Nesime Tatbul, Tae~Jun Lee, Stan Zdonik, Mejbah Alam and Justin Gottschlich,
  \emph{{Precision and recall for time series}}. In \emph{Proceedings of
  Advances in Neural Information Processing Systems}, Volume~31, 2018.

\bibitem[{Ulyanov} 2016]{Ulyanov2016}
Dmitry {Ulyanov}, Andrea {Vedaldi} and Victor {Lempitsky}, \emph{{Instance
  normalization: The missing ingredient for fast stylization}}. \emph{ArXiv
  e-prints}, 2016.

\bibitem[Wu 2020]{Wu2020}
Yuxin Wu and Kaiming He, \emph{{Group normalization}}. \emph{Int J Comput
  Vision}, Volume 128, Pages 742--755, 2020.

\bibitem[Yu 2020]{Yu2020}
Jiaqian Yu and Matthew~B. Blaschko, \emph{{The Lovasz hinge: A novel convex
  surrogate for submodular losses}}. \emph{IEEE Trans Pattern Anal Mach
  Intell}, Volume~42, Pages 735--748, 2020.

\bibitem[Zuo 2021]{Zuo2021}
Qiang Zuo, Songyu Chen and Zhifang Wang, \emph{{R2AU-Net: Attention recurrent
  residual convolutional neural network for multimodal medical image
  segmentation}}. \emph{Secur Commun Netw}, Pages 1--10, 2021.

\end{thebibliography}
\end{singlespace}
\end{document}